\PassOptionsToPackage{numbers,sort&compress}{natbib}
\documentclass[12pt]{elsarticle}
\usepackage[margin=2.5cm]{geometry}
\usepackage{enumitem}
\newlist{inlinelist}{enumerate*}{1}
\setlist*[inlinelist,1]{%
  label=(\roman*),
}
\usepackage{soul}
\usepackage{hhline}
\usepackage{multirow}
\usepackage{xcolor}
\usepackage{graphicx}
\usepackage{upgreek}
\usepackage{amssymb}
\usepackage{amsfonts,amsthm,bm,amsmath} % Math packages
\usepackage{appendix}
\usepackage{times}
\usepackage{caption}
\usepackage{subfig}

\usepackage{multirow}
\usepackage{xcolor}
\usepackage[linesnumbered,ruled,vlined]{algorithm2e}
\usepackage{tablefootnote}
\usepackage{comment}
\usepackage{bm}
\usepackage{tabularx}
\usepackage{caption}
\usepackage{array}
\usepackage{dsfont}

\SetCommentSty{mycommfont}

\SetKwInput{KwInput}{Input}                % Set the Input
\SetKwInput{KwOutput}{Output}              % set the Output

\usepackage[pagewise]{lineno}

\usepackage[colorlinks]{hyperref} 
\hypersetup{ 
    colorlinks=true,       % false: boxed links; true: colored links
    linkcolor=red,          % color of internal links
    citecolor=blue,        % color of links to bibliography
    filecolor=magenta,      % color of file links
    urlcolor=cyan           % color of external links
}

\newcommand{\fref}[1]{Fig.~\ref{#1}}

\newcommand{\eref}[1]{Eq.~(\ref{#1})}

\newcommand{\sref}[1]{Section~\ref{#1}}
\newcommand{\tref}[1]{Table~\ref{#1}}
\setcitestyle{square}

\usepackage{titlesec}
\usepackage[capitalise]{cleveref}

\titleformat{\paragraph}[hang]{\normalfont\itshape}{\theparagraph}{1em}{}
\titlespacing*{\paragraph}{0pt}{1ex plus 0.1ex minus 0.1ex}{0.5ex plus 0.1ex minus 0.1ex}

\makeatletter
\def\ps@pprintTitle{%
 \let\@oddhead\@empty
 \let\@evenhead\@empty
 \def\@oddfoot{\reset@font\hfil\thepage\hfil}%
 \let\@evenfoot\@oddfoot}
\makeatother

\begin{document}

\begin{frontmatter}

\title{Nonlinear Inverse Design of Mechanical Multi-Material Metamaterials Enabled by  Video Denoising Diffusion and Structure Identifier}
\author[]{Jaewan Park$^{1,2}$}
\author[]{Shashank Kushwaha$^{1}$}
\author[]{Junyan He$^{1}$}
\author[]{Seid Koric$^{1,2}$
\corref{mycorrespondingauthor}}
\cortext[mycorrespondingauthor]{Corresponding author}
\ead{koric@illinois.edu}
\author[]{Qibang Liu$^{2,3}$}
\author[]{Iwona Jasiuk$^1$}
\author[]{Diab Abueidda$^{2,4}$\corref{mycorrespondingauthor}}
%\cortext[mycorrespondingauthor2]{Corresponding author}
\ead{abueidd2@illinois.edu}
\address{$^1$ Department of Mechanical Science and Engineering, University of Illinois at Urbana-Champaign, Urbana, IL, USA \\
$^2$ National Center for Supercomputing Applications, University of Illinois at Urbana-Champaign, Urbana, IL, USA \\

$^3$ Department of Industrial and Manufacturing Systems Engineering, 
Kansas State University,
Manhattan, KS, USA \\
$^4$ Civil and Urban Engineering Department, New York University Abu Dhabi, United Arab Emirates\\
}

\begin{abstract}

Metamaterials, synthetic materials with customized properties, have emerged as a promising field due to advancements in additive manufacturing. These materials derive unique mechanical properties from their internal lattice structures, which are often composed of multiple materials that repeat geometric patterns. While traditional inverse design approaches have shown potential, they struggle to map nonlinear material behavior to multiple possible structural configurations. This paper presents a novel framework leveraging video diffusion models, a type of generative artificial Intelligence (AI), for inverse multi-material design based on nonlinear stress-strain responses. Our approach consists of two key components: (1) a fields generator using a video diffusion model to create solution fields based on target nonlinear stress-strain responses, and (2) a structure identifier employing two UNet models to determine the corresponding multi-material 2D design.  By incorporating multiple materials, plasticity, and large deformation, our innovative design method allows for enhanced control over the highly nonlinear mechanical behavior of metamaterials commonly seen in real-world applications. It offers a promising solution for generating next-generation metamaterials with finely tuned mechanical characteristics.

\end{abstract}


\end{frontmatter}

%\linenumbers

\section{Introduction}
\label{sec:intro}

The concept of producing artificial materials with customized characteristics has become increasingly popular in many engineering fields since additive manufacturing allows for the fabrication of architected materials, also known as metamaterials, at various scales and sizes \cite{nazir2023multi}. Unique mechanical properties of metamaterials, such as how they bend, stretch, compress, or respond to forces, come from the material’s internal structure, often made up of repeating geometric patterns of multiple conventional materials. Opposite to the restricted range of natural bulk materials, engineers and designers now have the opportunity to explore the significantly broadened design and property possibilities of metamaterials. These materials have been specifically developed to accomplish mechanical properties and behaviors that were previously unattainable, such as in impact energy absorption in automotive and aerospace industries \citep{siddique2022lessons}, soft robotics and actuators that can undergo large and irreversible deformations in response to external loadings \citep{pan2020programmable}, and bio-inspired materials for that mimic the nonlinear mechanical properties and behavior of biological tissues for use in tissue engineering or medical implants and prosthetic \citep{dogan20203d}. Recent advancements in multi-material additive manufacturing have further expanded the possibilities for designing and fabricating complex lattice structures \cite{nazir2023multi}. Such capability allows for more precise control over the mechanical properties and behavior of metamaterials, opening up new avenues for innovative designs and applications.

Historically, the forward approaches \citep{zheng2018minimal, wang2021structured, scarpa_2000, VISWANATH2023} were used to systematically modify the design parameters of complex and architected materials until the measured or simulated properties meet specifically defined design requirements. These methods typically need significant prior knowledge of experienced designers, and they are burdened by computationally demanding simulations of iterative experiments, especially for materials exhibiting intricate, nonlinear characteristics. On the other hand, topology optimization (TO) methods, including gradient-based algorithms and genetic algorithms \citep{xia2014concurrent, takezawa2017design, diab_topo_nme_2021, zeng2023inverse}, and traditional non-generative machine learning-based design approaches \citep{bessa2019bayesian, kumar2020inverse, wilt2020accelerating, kollmann2020deep, abueidda2020topology, kushwaha2024advanced}, as reviewed in \citep{zheng2023_dl_review}, have shown potential for achieving desired property values. However, these methods face several challenges when dealing with complex, nonlinear material behaviors. While TO can be extended to various types of nonlinearities \citep{james2015topology}, such extensions are mathematically rigorous, and their implementation is prone to errors. Additionally, TO methods often struggle with getting trapped in local minima, particularly for highly nonlinear problems. Furthermore, the non-unique mapping between mechanical responses and structural designs, inherent in many metamaterial design problems, poses a significant challenge for these approaches. 

Generative artificial intelligence models represent a category of artificial intelligence systems engineered to produce novel data that emulates patterns found in existing data. Generative models differ from traditional AI models, primarily focused on recognizing patterns or making predictions based on input data. Instead, generative models seek to produce novel outputs that mimic the training data's characteristics, making them attractive for nonlinear inverse design \citep{parrott2023multidisciplinary, parrott2023multi}.
Mao et al. \citep{mao2020designing} used a generative adversarial network (GAN) to design architected materials while Zheng et al. \citep{ZHENG2021} proposed an inverse design method for auxetic metamaterials using a conditional generative adversarial network cGAN. 
\citet{ma2019probabilistic} and \citet{WANG2020} used a conditional variational autoencoder (VAE) to generate a metamaterial structure with specific mechanical properties.
Recently, Ha et al. \citep{ha2023rapid} implemented a generative machine learning (ML) pipeline composed of inverse prediction and forward validation modules. 

Among generative AI models, diffusion models \cite{ho2020denoising} have shown exceptional achievements in several domains of AI,
notably in generating images and videos. Diffusion models are ML methods that gradually transform random noise into meaningful data,
like images or videos. Starting with a clear image or a video, the model first adds random noise until it becomes unrecognizable and pure noise.
Once the model learns how the noise is added, it reverses the process by gradually denoising it step by step,
transforming the noise back into something recognizable and comparing it with its original image or video target.
By repeating this training process with many data samples, the model eventually learns to generate new images or
videos from random noise data it has never seen before. The newly generated images or videos look realistic and match the patterns
learned from the original data. To create images or videos on specific conditional class, \citet{dhariwal2021diffusion} 
proposed a classifier guidance method that trains a classifier on noise images and uses the gradient of the classifier to guide the 
sampling toward the target class. Later, \citet{ho2022classifier} proposed a classifier-free guidance method that uses the
scores from a conditional diffusion and an unconditional diffusion model,
allowing the model to learn to generate fields both conditionally and unconditionally. 
Such diffusion models have been widely used in current popular AI applications, 
such as DALL-E-2  \cite{ramesh2022hierarchical}, DALL-E-3 \cite{betker2023improving}, and 
Sora \cite{videoworldsimulators2024,liu2024sora} developed by OpenAI, Imagen Video \cite{NEURIPS2022_ec795aea, ho2022imagen} by Google Research Brain Team, Stable Diffusion \cite{rombach2022high,podell2023sdxl,sauer2024fast} by Stability AI, Midjourney by Minjourney Inc,
and NAI diffusion by NovelAI.
Beyond image or video synthesis in these tools, application of conditional diffusion models has expanded 
recently to encompass disciplines such as design for material science, including molecular design \cite{weiss2023guided,liu2024inverse,bao2022equivariant},
single metamaterial architecture design \citep{bastek2023inverse, yang2024guided,jadhav2024generative, vlassis2023denoising}, 
superconductors design \citep{wines2023inverse, zhong2024high} and many others, where they reveal promise for addressing highly challenging and
often unsolvable inverse design problems. 

This paper presents a novel framework that leverages a video diffusion model that performs inverse multi-material design. Our approach is specifically tailored to generate designs with multiple materials that satisfy a given nonlinear stress-strain response. The application of diffusion models to video data incorporates temporal dynamics, enabling the model to accurately represent the dynamic changes that occur over time, similar to the evolution of deformation paths in nonlinear mechanical quasi-static behavior. The capacity to represent both spatial and temporal relationships renders video diffusion models suitable for challenges involving the evolution of system changes over time, such as in nonlinear deformation under stress. Bastek and Kochmann \citep{bastek2023inverse} used this concept, inspired by \cite{NEURIPS2022_ec795aea, ho2022imagen}, to design periodic stochastic cellular structures under compression loading in the large-strain hyper-elastic regime, including buckling and contact. However, their landmark work involved a single material, which enabled a simple mapping methodology to predict material distribution from the predicted solution fields. To the best of our knowledge, no previous research work has devised a general diffusion model framework to predict metamaterial designs with multiple materials. Such problem is very important since designers can further fine-tune the metamaterial's overall behavior by using different materials, each with unique mechanical, thermal, or electrical properties. For example, one material might provide stiffness while another offers compliance, allowing for a balance between strength and deformability. In addition, we included plastic deformation, an irreversible behavior common to many materials under large deformation that exceeds yielding.  Conversely, this also introduced additional challenges to our diffusion model in learning deformation path-dependent highly nonlinear material behavior.

\section{Methods}
\label{sec:methods}

\subsection{Training data generation}
\label{sec:training_data_generation}

\begin{figure}[h!] 
    \centering
    \subfloat[Examples of multi-material designs created using Voronoi polygons]{
         \includegraphics[trim={0cm 0cm 0cm 0cm},clip,width=0.95\textwidth]{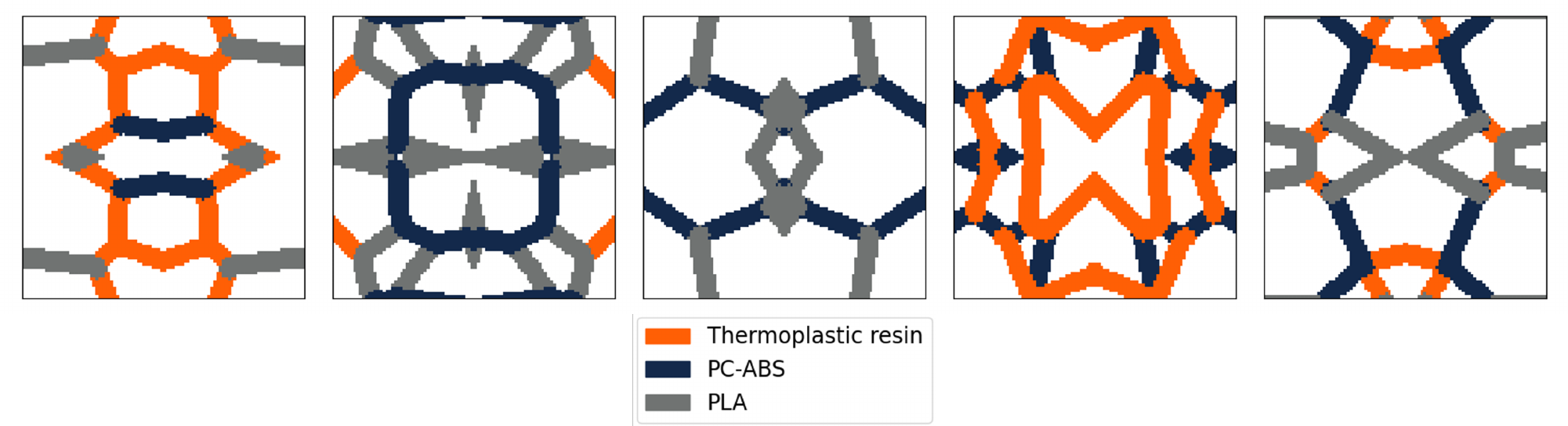}
         \label{Voronoi_designs}
     }
    \\
    \subfloat[Finite element setting to obtain field values and stress-strain curve of the whole structure.]{
         \includegraphics[trim={0cm 0cm 0cm 0cm},clip,width=0.6\textwidth]{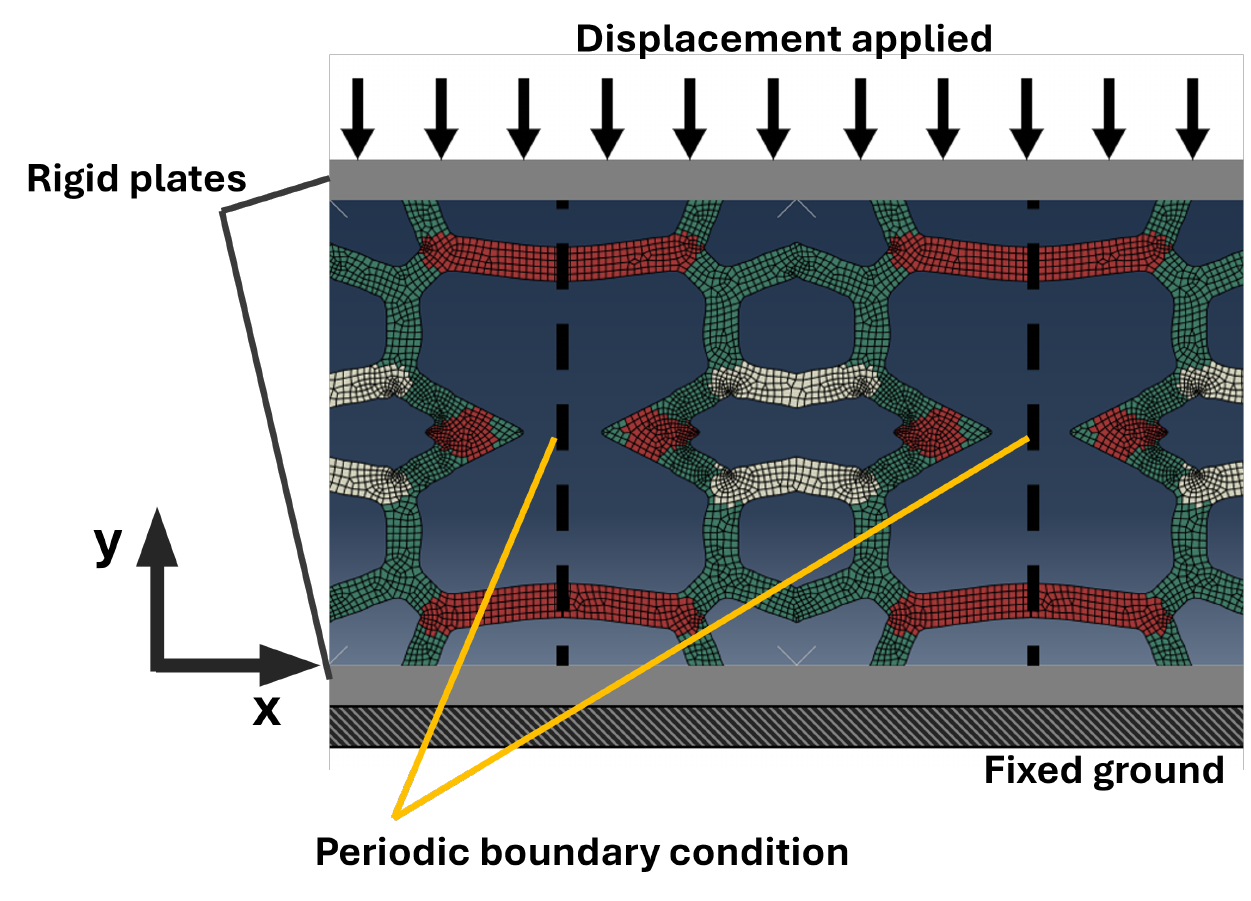}
         \label{Abaqus_setting}
     }
    \caption{Designs examples and FEA setting of the problem}
\label{Design_and_setting}
\end{figure}

We generate truss-like multi-material metamaterials based on random Voronoi tessellations \citep{voronoi1908nouvelles}. The initial step involves generating random points in a 2D space and using these points to create finite Voronoi polygons through the Voronoi function in the Python library \textit{scipy.spatial}. The method constructed Voronoi diagrams, which are then used to extract polygon edges representing the boundaries between regions. Each of these edges is randomly assigned one of three material types, forming the basis of the material distribution. Following this, the polygon edges are thickened. Therefore, each edge becomes a truss-like feature in the design. For simplicity of representing the designs and using the design information as inputs to the diffusion model, the thickened design is then projected to a structured grid to be stored as a 2D array of shape 48$\times$48, where each pixel stores the material ID (0 for void, 1-3 for the three different materials). To construct a symmetric and periodic design, each instantiation is considered as one-quarter of the full design, and the full design is recovered from sequential mirroring along two in-plane directions. The use of random Voronoi tessellations leads to metamaterials with varying relative density, further leading to design space with a wide range of stress-strain responses. Furthermore, this pixel-based representation makes it ideal for use in finite element simulations. Thus, this method provides a flexible approach for generating non-regular, multi-material distributions with varying degrees of complexity, applicable to broad engineering applications. Selected example designs are shown in \fref{Voronoi_designs}.

Once the designs are generated, nonlinear implicit dynamic analysis is performed using general purpose finite element analysis (FEA) code \cite{Abaqus2024}, solving combined nonlinearities originating from large deformation and plastic material response. As shown in \fref{Abaqus_setting}, designs are placed between two rigid plates, and a quasi-static compression load is applied by moving the top plate and keeping the bottom plate fixed. The structures are compressed to 15\% nominal strain. Periodic boundary conditions are enforced in the left and right vertical edges of the structure to obtain a periodic response similar to that observed in long sandwich structure configuration. The materials used in the study are defined using an elastoplastic material model and are commonly used in additive manufacturing applications. The material properties taken from the literature are close to the following materials: PC-ABS \citep{kushwaha2023designing}, thermoplastic acetal homopolymer resin \citep{jin2020guided}, and PLA \citep{johnston2021analysis}. In \fref{Voronoi_designs} and throughout this paper, orange represents thermoplastic resin, navy represents PC-ABS, and gray represents PLA. Frictional self-contact was defined with a friction coefficient value of 0.4, and structure nodes at the top and bottom plates were allowed to slip horizontally relative to the horizontal rigid plates. The generated mesh was compatible with the periodic boundary conditions, and four-node bilinear elements (CPE4R) with hourglass control and reduced integration were used.  Plain-strain condition is assumed to represent a design extruded in the out-of-plane dimension. Hence, out-of-plane buckling is not considered in this study. The workflow to smooth the design boundaries and generate a conformal mesh from a 2D array of design is adopted from the work of \citep{bastek2023inverse}.

The workflow defined above was used to generate 3314 simulations. All simulations involving physics and finite element analysis were conducted on Delta's High-Performance Computing (HPC) system hosted at the National Center for Supercomputing Applications (NCSA). The CPU nodes of Delta are equipped with AMD EPYC 7763 Milan CPU cores. On average, each simulation required approximately 232 seconds to complete. Using high throughput computing allowed running up to 12 different FE simulations simultaneously. Hence, the total time required for data generation was 18 hours, corresponding to 216 CPU hours. Four output fields are extracted from the result files at 11 equidistant strain increments up to 15\% nominal strain. They are the von Mises stress $\sigma_{vM}$, the axial stress $\sigma_{yy}$, the strain energy density $g_{se}=\int^{\epsilon_f}_0 \bm{\sigma} : \bm{\epsilon} \, d \bm{\epsilon} $ and the lateral displacement $u_x$. $\sigma_{vM}$ is included since it is a commonly used scalar measure for the material stress state, and it is closely related to material yielding. $\sigma_{yy}$ is included to complement $\sigma_{vM}$ to indicate whether the material point is under tension or compression. $g_{se}$ is included to further differentiate different materials in different parts of the design. As for the displacements, only $u_x$ is included since the vertical displacement $u_y$ is fixed at the bottom rigid plate locations and does not contain as much information as the former. For simplicity with handling the different FE meshes for different designs, the nodal output fields at the finite element mesh nodes are interpolated to a constant 96$\times$96 pixel grid for all cases. The interpolated fields are stored as 3D arrays each of shape 96$\times$96$\times$4.

\subsection{Neural network framework}
\begin{figure}[h!] 
    \centering
         \includegraphics[width=0.97\textwidth]{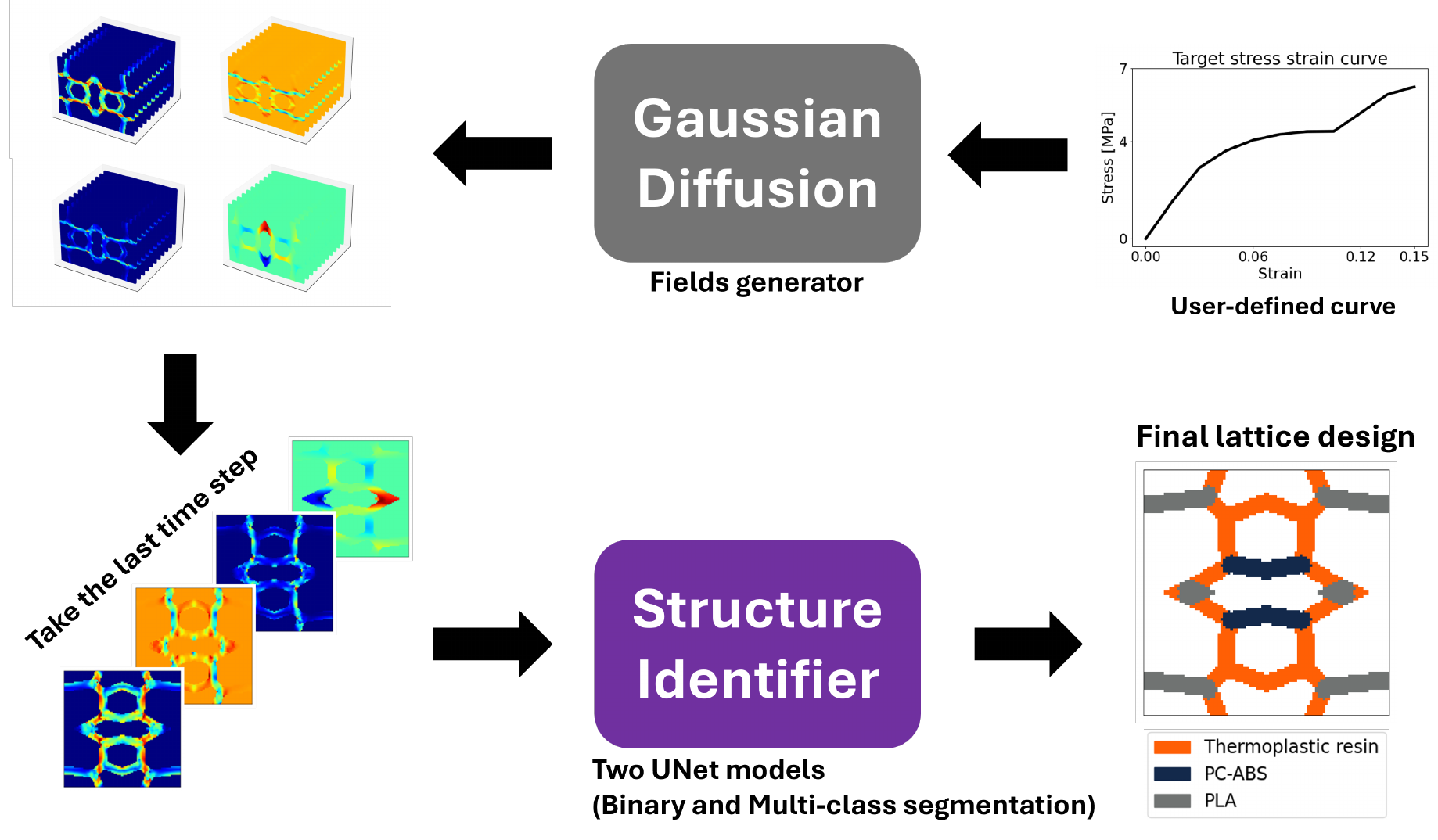}
    \caption{Whole framework on generating a multi-material lattice design from a single stress-strain curve.}
    \label{Whole_framework}
\end{figure}

This section provides an overview of the neural network (NN) models employed in the proposed framework to generate 2D multi-material designs. As illustrated comprehensively in \fref{Whole_framework}, the framework consists of two primary components: the Gaussian diffusion model (i.e., a fields generator) that learns to generate field contours based on a target force-displacement curve, and the structure identifier (i.e., two UNet models) that identifies the corresponding multi-material 2D design from the generated fields. Each component plays a crucial role in creating novel designs through generative AI and image segmentation.

\subsubsection{Diffusion model as a fields generator}
In this work, we adopted a classifier-free guided diffusion model from \citep{bastek2023inverse} to generate four mechanical solution fields (i.e., $\sigma_{vM}$, $\sigma_{yy}$, $g_{se}$ and $u_x$), based on the user-provided strain-stress curve. This section provides a brief review of the
denoising diffusion probabilistic model \cite{ho2020denoising} and
the classifier-free guidance method \citep{ho2022classifier} for completeness.

Given a sample $\mathbf{x}_0$ from a prior data distribution $\mathbf{x}_0 \sim q(\mathbf{x})$,
the forward diffusion process of the denoising diffusion probabilistic model adds a small amount of Gaussian
noise to the sample over $\mathbf{T}$ steps. This results in a sequence of samples $\mathbf{x}_1, \mathbf{x}_2, \ldots, \mathbf{x}_T$.
Each step in the diffusion process is controlled by a variance schedule $\left\{\beta_t \in(0,1)\right\}_{t=1}^T$,
\begin{subequations}\label{eq:diffusing}
    \begin{align}
        & q\left(\mathbf{x}_t|\mathbf{x}_{t-1}\right)=\mathcal{N}\left(\mathbf{x}_t;
        \sqrt{1-\beta_t} \mathbf{x}_{t-1}, \beta_t \mathbf{I}\right), \\
        & q\left(\mathbf{x}_{1: T}|\mathbf{x}_0\right)=\prod_{t=1}^T q\left(\mathbf{x}_t|\mathbf{x}_{t-1}\right).
    \end{align}
\end{subequations}
The sample $\mathbf{x}_0$ gradually loses its features and eventually becomes an isotropic Gaussian distribution for large $T$.
This diffusion process has the elegant property that we can sample $\mathbf{x}_t$ at any time step $t$ using the reparametrization trick,
\begin{subequations}\label{eq:sampling}
    \begin{align}
    &\mathbf{x}_t = \sqrt{\bar{\alpha}_t} \mathbf{x}_0 + \sqrt{1 - \bar{\alpha}_t} \epsilon_t, \\
    &q\left(\mathbf{x}_t|\mathbf{x}_0\right) = \mathcal{N}\left(\mathbf{x}_t ; \sqrt{\bar{\alpha}_t} \mathbf{x}_0, \left(1 - \bar{\alpha}_t\right) \mathbf{I}\right),
    \end{align}
\end{subequations}
where $\alpha_t = 1 - \beta_t$, $\bar{\alpha}_t = \prod_{i=1}^t \alpha_i$, and $\epsilon_t \sim \mathcal{N}(0, \mathbf{I})$.

By reversing the diffusion process, we can sample from $q(\mathbf{x}_{t-1}| \mathbf{x}_t, \mathbf{x}_0)$ and ultimately
 generate the true sample $\mathbf{x}_0$ from Gaussian noise $\mathbf{x}_T \sim \mathcal{N}(0, \mathbf{I})$.
 Using Bayesian rule, the reverse posterior distribution conditioned on $\mathbf{x}_0$ is traced,
\begin{equation}
    q\left(\mathbf{x}_{t-1}|\mathbf{x}_t, \mathbf{x}_0\right) = \mathcal{N}\left(\mathbf{x}_{t-1} ; \tilde{\boldsymbol{\mu}}_t\left(\mathbf{x}_t, \mathbf{x}_0\right), \tilde{\beta}_t \mathbf{I}\right),
\end{equation}
where $\tilde{\boldsymbol{\mu}}_t\left(\mathbf{x}_t, \mathbf{x}_0\right) = \frac{1}{\sqrt{\alpha_t}}\left(\mathbf{x}_t - \frac{1 - \alpha_t}{\sqrt{1 - \bar{\alpha}_t}} \epsilon_t\right)$
and $\tilde{\beta}_t = \frac{1 - \bar{\alpha}_{t-1}}{1 - \bar{\alpha}_t} \beta_t$.

In a real generative process, we cannot directly evaluate $q\left(\mathbf{x}_{t-1}|\mathbf{x}_t, \mathbf{x}_0\right)$
because it requires the entire training samples and the diffusion noise dataset.
Instead, we propose a posterior distribution $p_\theta\left(\mathbf{x}_{t-1}|\mathbf{x}_t\right)$ to
approximate the true posterior distribution $q\left(\mathbf{x}_{t-1}|\mathbf{x}_t, \mathbf{x}_0\right)$,
\begin{equation}
    p_\theta\left(\mathbf{x}_{t-1}|\mathbf{x}_t\right) = \mathcal{N}\left(\mathbf{x}_{t-1} ; \boldsymbol{\mu}_\theta\left(\mathbf{x}_t, t\right), \mathbf{\Sigma}_\theta\left(\mathbf{x}_t, t\right)\right),
\end{equation}
The reverse denoising process is then controlled by,
\begin{equation}
    p_\theta\left(\mathbf{x}_{0:T}\right) = p\left(\mathbf{x}_T\right) \prod_{t=1}^T p_\theta\left(\mathbf{x}_{t-1}|\mathbf{x}_t\right),
\end{equation}
where $\mathbf{\Sigma}_\theta\left(\mathbf{x}_t, t\right)$ is the same as $\tilde{\beta}_t \mathbf{I}$,
and $\boldsymbol{\mu}_\theta\left(\mathbf{x}_t, t\right)$ has the same form as
$\tilde{\boldsymbol{\mu}}_t\left(\mathbf{x}_t, \mathbf{x}_0\right)$ but with learnable parameters $\theta$,
\begin{equation}\label{eq:approx_mu}
    \boldsymbol{\mu}_\theta\left(\mathbf{x}_t, t\right) = \frac{1}{\sqrt{\alpha_t}}\left(\mathbf{x}_t - \frac{1 - \alpha_t}{\sqrt{1 - \bar{\alpha}_t}} \epsilon_\theta\left(\mathbf{x}_t, t\right)\right).
\end{equation}
The optimization objective of the denoising diffusion probabilistic model is to
maximize the log-likelihood $\log p_\theta\left(\mathbf{x}_{0:T}\right)$, which results in the following loss function,
\begin{equation}
    L = \left|\epsilon_\theta\left(\mathbf{x}_t, t\right) - \epsilon_t \left(\mathbf{x}_t, t\right)\right|.
\end{equation}
Training a NN $\epsilon_\theta\left(\mathbf{x}_t, t\right)$ and minimizing the loss function $L$ 
leads to $\boldsymbol{\mu}_\theta\left(\mathbf{x}_t, t\right) \approx \tilde{\boldsymbol{\mu}}_t\left(\mathbf{x}_t, \mathbf{x}_0\right)$,
thus making the proposed posterior distribution $p_\theta\left(\mathbf{x}_{t-1}|\mathbf{x}_t\right)$ close
to the true posterior distribution $q\left(\mathbf{x}_{t-1}|\mathbf{x}_t, \mathbf{x}_0\right)$.

The above reverse generative process is random and not controlled by any specific target.
In specific design tasks, we aim to generate fields that represent the target, such as the force-displacement curve,
which requires training the NN $\epsilon_\theta\left(\mathbf{x}_t, t\right)$ with conditional information.
To incorporate the condition information $\mathbf{y}$ into the diffusion process,
\citet{ho2022classifier} proposed a classifier-free guidance method without an explicit classifier \cite{dhariwal2021diffusion},
incorporating the scores from a conditional diffusion model $p_\theta (\mathbf{x}|\mathbf{c})$
and an unconditional diffusion model $p_\theta (\mathbf{x})$. 
The noise estimators $\epsilon_\theta\left(\mathbf{x}_t, t\right)$ of $p_\theta (\mathbf{x})$
and $\epsilon_\theta\left(\mathbf{x}_t, t, \mathbf{c}=\mathbf{y}\right)$ of $p_\theta (\mathbf{x}|\mathbf{c})$
are trained in a single NN $\epsilon_\theta\left(\mathbf{x}_t, t, \mathbf{c}\right)$.
Here, $\epsilon_\theta\left(\mathbf{x}_t, t, \mathbf{c}=\mathbf{y}\right)$ is trained with paired data $(\mathbf{x}, \mathbf{c}=\mathbf{y})$,
and $\epsilon_\theta\left(\mathbf{x}_t, t\right)$ is trained with data $\mathbf{x}$ only, i.e., $(\mathbf{x}, \mathbf{c}=\emptyset)$. 
During the training process, the condition information $\mathbf{c}$ is randomly set as $\mathbf{c}=\emptyset$
or $\mathbf{c}=\mathbf{y}$ for sample $(\mathbf{x}, \mathbf{y})$ at different time steps $t$, 
allowing the model to learn to generate fields both conditionally and unconditionally.
During the reverse inference process, $\epsilon_\theta$ in \cref{eq:approx_mu} is replaced by the
linear summation of conditional and unconditional noise estimators,
\begin{equation}
    \epsilon_\theta\left(\mathbf{x}_t, t, \mathbf{c}\right) = (1 + w)\epsilon_\theta\left(\mathbf{x}_t, t, \mathbf{c}=\mathbf{y}\right) - w \epsilon_\theta\left(\mathbf{x}_t, t, \mathbf{c}=\emptyset\right),
\end{equation}
where $w \ge 0$ is the guidance weight.

\begin{figure}[h!] 
    \centering
    \subfloat[Down and Up blocks used in the Gaussian diffusion model]{
         \includegraphics[trim={0cm 0cm 0cm 0cm},clip,width=0.95\textwidth]{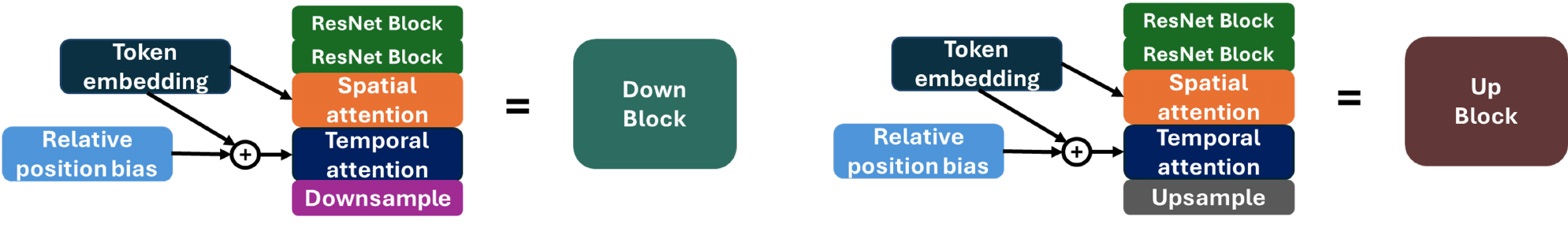}
         \label{Diffusion_blocks}
     }
    \\
    \subfloat[Training phase of the Gaussian diffusion model. 3D UNet serves a key algorithm for predicting noise added to each step in the diffusion process.]{
         \includegraphics[trim={0cm 0cm 0cm 0cm},clip,width=0.95\textwidth]{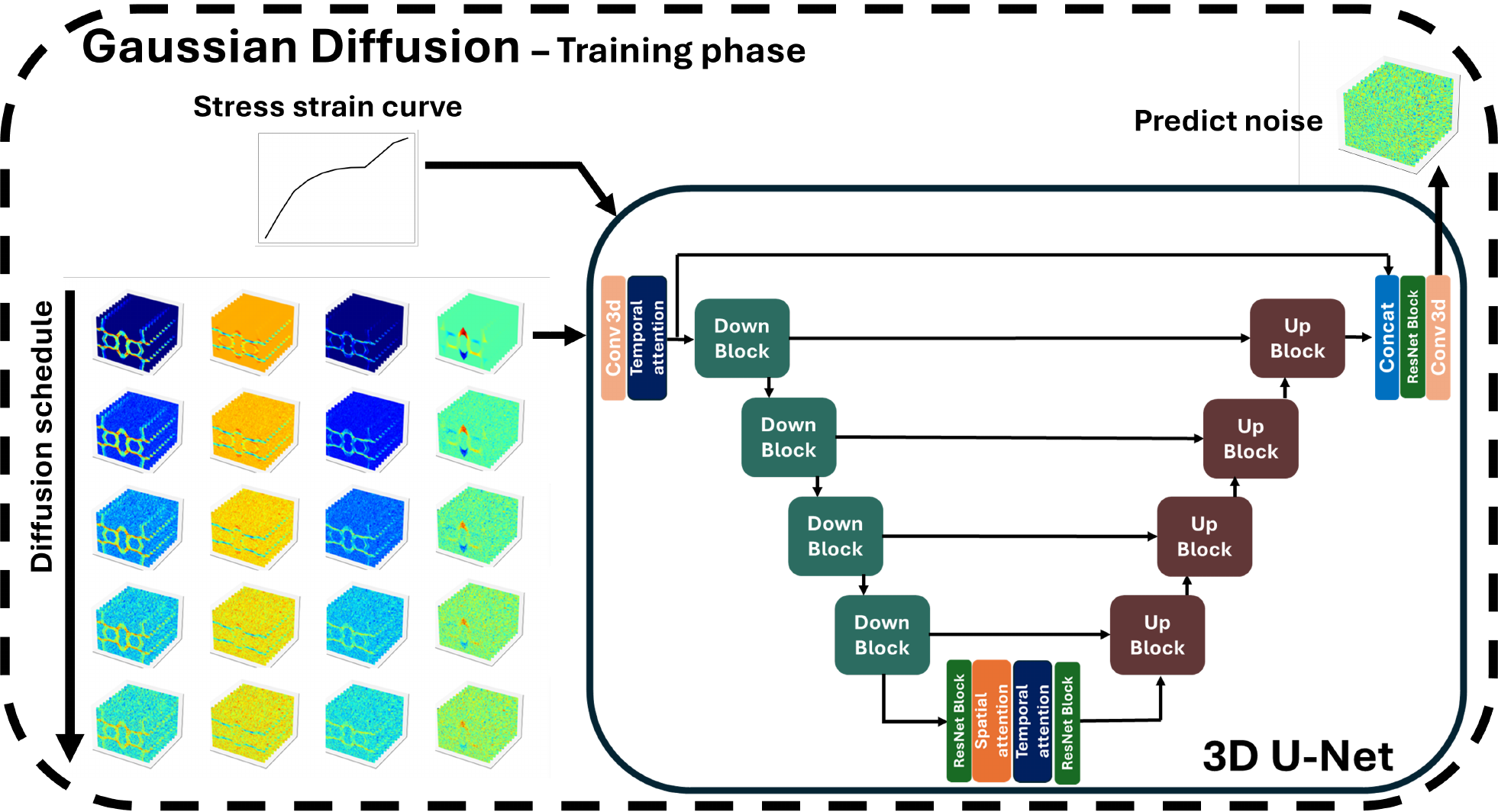}
         \label{Diffusion_structure_training}
     }
     \\
    \subfloat[Sampling phase of the Gaussian diffusion model. Trained 3D UNet takes the random Gaussian noise and user-defined target curve as an input, then serves role of a denoising algorithm to generate fields from the given random noise.]{
         \includegraphics[trim={0cm 0cm 0cm 0cm},clip,width=0.95\textwidth]{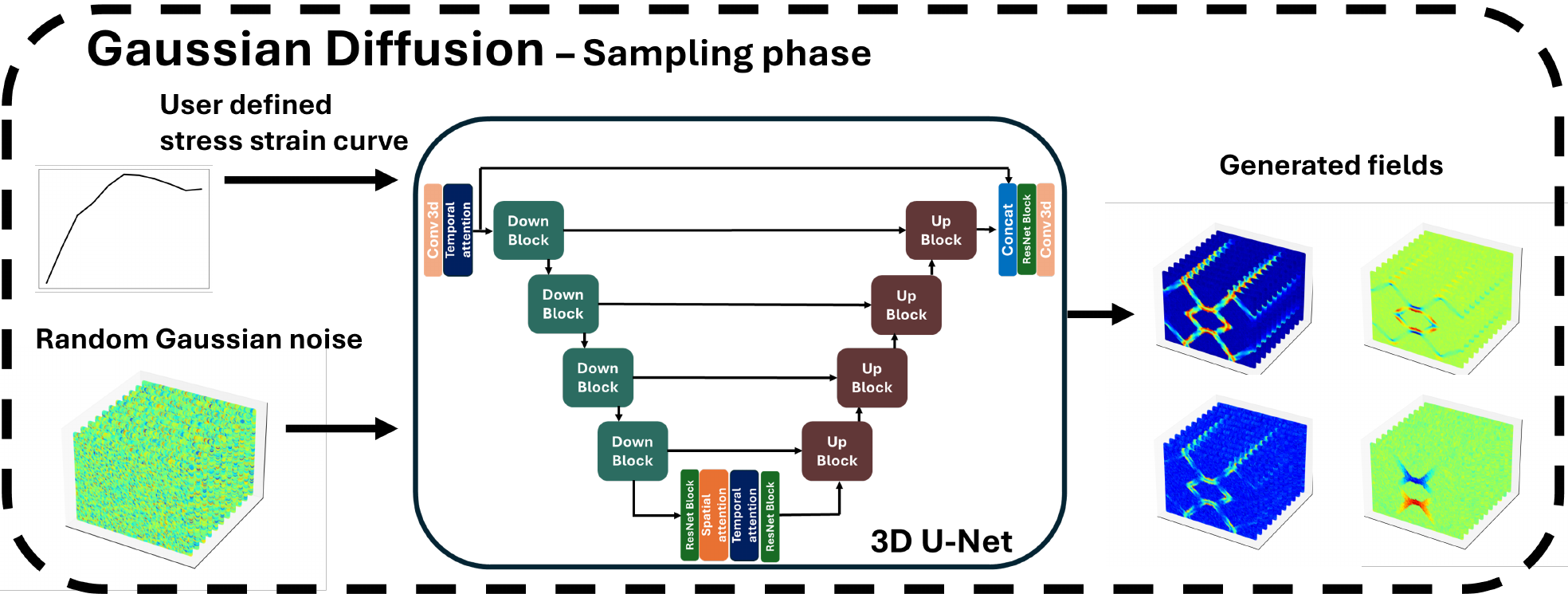}
         \label{Diffusion_structure_sampling}
     }
    \caption{Architecture of the Gaussian diffusion model: Down and Up blocks used in the 3D UNet, and overall structure of the diffusion process with 3D UNet as the core noise prediction algorithm}
\label{GD_block_and_structure}
\end{figure}

The classier-free guided diffusion model predicts these solution fields by iteratively denoising random noise inputs, resulting in multiple sets of 2D field values, representing field values for each time step that serves as the basis for subsequent design structure identification. A 3D UNet architecture was integrated with the Gaussian Diffusion model, serving as the core model for the generative process due to its ability to handle multiple 2D images as a time sequence. Given the spatial and temporal information embedded in the data, spatial linear self-attention and temporal self-attention mechanisms were incorporated with ResNet blocks within each encoder and decoder block of the 3D UNet. Each attention block contains 8 attention heads. The network structure includes four upsampling and downsampling encoder-decoder blocks, resulting in a total depth of five, including the input layer. Schematics of the encoding and decoding blocks used in the model are shown in \fref{Diffusion_blocks}. To handle temporal strain steps, which consist of 11 stress values, a linear layer was employed to convert these steps into a token embedding vector.

\fref{Diffusion_structure_training} illustrates the training phase of the model. The structure is based on a forward diffusion process, where a given design is gradually corrupted by adding Gaussian noise over a series of time steps. A 3D UNet plays a crucial role in predicting the noise added at each step. The model learns to reverse this process, iteratively denoising the input to recover the original design. The sampling phase, depicted in \fref{Diffusion_structure_sampling}, begins with pure noise and processes the reverse Markov chain to achieve our goal: obtaining 2D field values at each time step. This process is governed by the same noise schedule used in training but applied in reverse order.

\subsubsection{Two UNets as a structure identifier}
\label{sec:Two_UNets}
Similar to the work of \cite{bastek2023inverse}, the diffusion model generates mechanical solution fields instead of the desired design directly. In addition, due to the stochastic nature of the diffusion process, these raw outputs may contain noise and other imperfections that need to be properly addressed to obtain a physically valid structure. A simple stress threshold is used in \cite{bastek2023inverse} to identify the void and solid regions for a single material. This would not work for a multi-material scenario such as the one considered in this work. Therefore, in this work, we proposed to use two consecutive UNets to post-process the fields generated by the diffusion model to obtain the final design: the first one performs a binary segmentation to identify solid and void regions, and the other one performs a multi-class segmentation to identify the material distribution within the solid region. This two-step structure identification process ensures that the final designs are coherent with the designs created initially during the training data generation phase and renders the structure identification process more explainable. 

\begin{figure}[h!] 
    \centering
    \subfloat[Encoder and decoder block structure for the UNets used]{
         \includegraphics[trim={0cm 0cm 0cm 0cm},clip,width=0.95\textwidth]{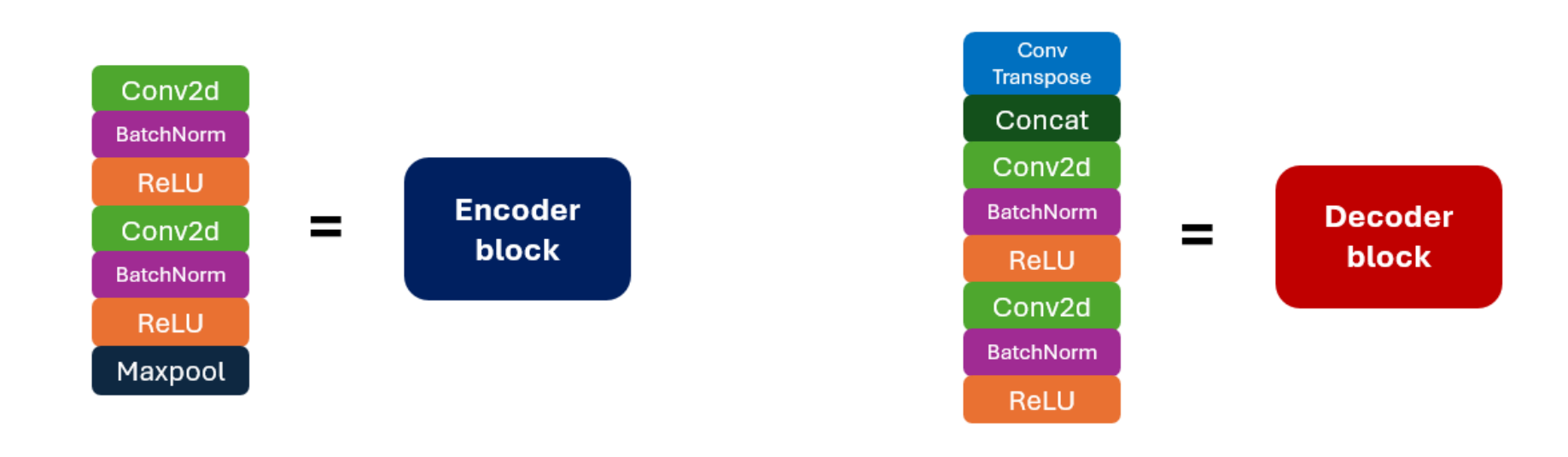}
         \label{Encoder_decoder_block}
     }
    \\
    \subfloat[Single UNet architecture used in the structure identifier. The $1^{st}$ and $2^{nd}$ UNets share this structure, differing only in the number of output classes.]{
         \includegraphics[trim={0cm 0cm 0cm 0cm},clip,width=0.95\textwidth]{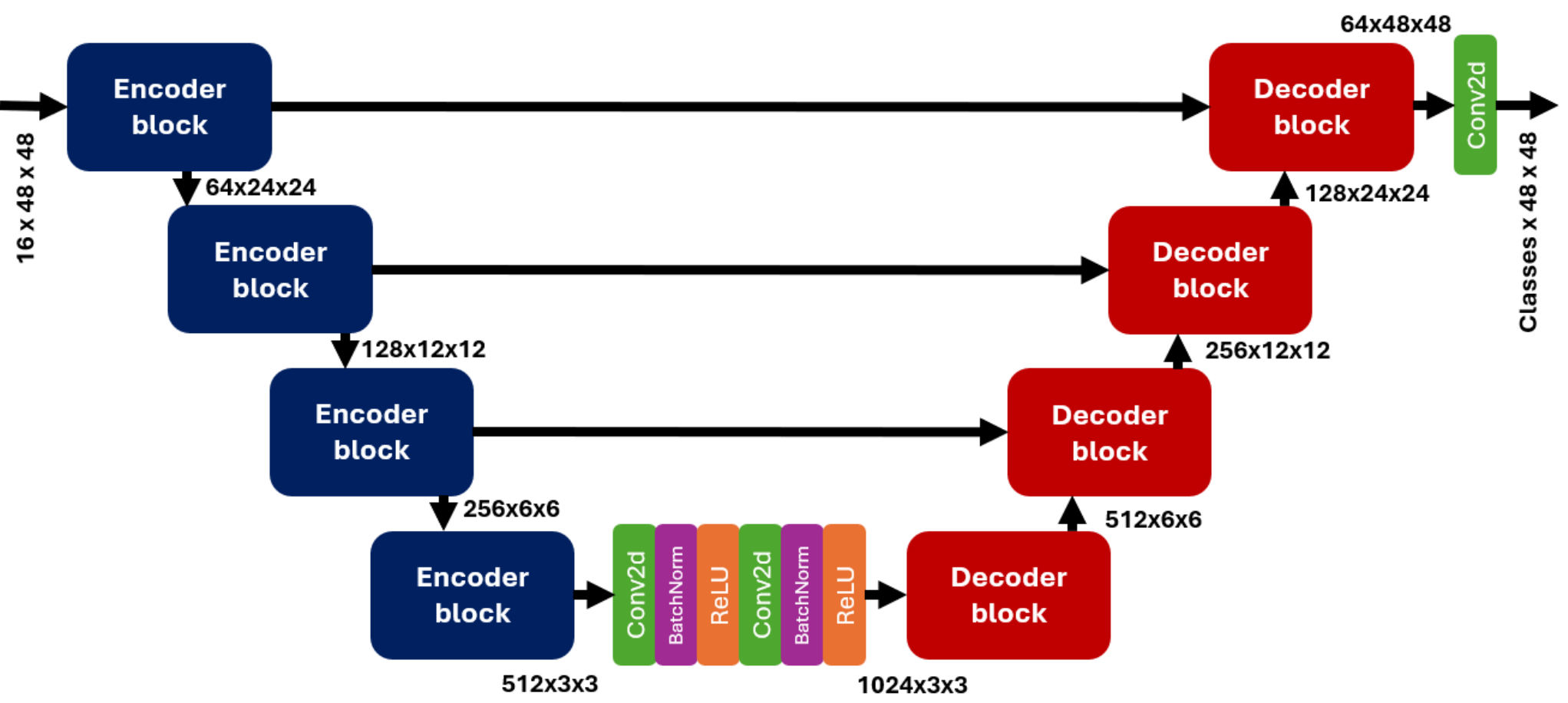}
         \label{UNet_structure}
     }
     
    \caption{Block structure and overall architecture of the UNet-based structure identifier}
    \label{UNet_block_and_structure}
\end{figure}

The first UNet is designed to perform a binary segmentation on the output fields from the diffusion model. We used the last time step of the Gaussian diffusion outputs from each channel as a training data, since the fields at this final step show the most distinct contrast with void regions. Then it (i.e., the mechanical fields obtained from simulations) was augmented by randomly selecting one in the following list: Random horizontal flip, Random vertical flip, Random rotation by 90 degrees, and Gaussian blur. With data augmentation, the training data was expanded 6 times of its original size to enhance generalizability. To enhance resistance of the model to the inherent noise in the diffusion-predicted field contours, random Gaussian noises were added to the training data field values to resemble the noisy diffusion-generated output, as the diffusion algorithm iteratively adds Gaussian noise in its training phase. Finally, to ensure symmetry in the design, each full-field data (of shape 96 $\times$ 96) was divided into four symmetric parts, four 48 $\times$ 48 images, and stacked along the channel axis. Each part was rotated and flipped to match the orientation of the upper-left slice when stacking. As a result, the channel axis became four times larger than before. The augmented-then-noise-enhanced dataset is used in the training of both UNet models in the structure identifier. The binary segmentation UNet employs an encoder-decoder structure that processes the input to distinguish between material and void regions. The encoder and decoder blocks are constructed out of multiple layers of computation as \fref{Encoder_decoder_block} and the binary cross entropy loss were used as a loss function. The output from this structure is a binary image with pixels of 48 $\times$ 48, where the material regions are identified against the void background. A schematic of the UNet architecture is shown in \fref{UNet_structure}.

After the initial binary segmentation to differentiate solid and void regions, a second UNet now undertakes the task of multi-class segmentation to obtain the final multi-material design. The training data pre-treated with data augmentation and added noise are used in the training of this second UNet, except that now the fields in the void region (predicted from the first UNet) are set to 0 before training. This additional treatment step ensures that the second UNet can easily segment out void pixels and concentrate more effectively on the material pixels to perform the material classification task. The architecture follows an encoder-decoder structure similar to that of the first UNet, but with modifications in the final layer to classify each pixel into specific material classes. Since there are 3 materials and a void to classify for each pixels, channel of 4 was set to be the output from the second UNet. The main difference from the first UNet lies in the loss function, where we use the DiceCE loss defined as:
\begin{equation}
\begin{aligned}
    \mathcal{L}_{\text {DiceCE }}=\mathcal{L}_{\text {Dice }}+\mathcal{L}_{\mathrm{CE}}, \\
    \mathcal{L}_{\text {Dice }}=\frac{1}{C} \sum_{c=1}^C\left(1-\frac{2 \sum_{i=1}^N y_{i, c} \sigma\left(z_{i, c}\right)+\epsilon}{\sum_{i=1}^N y_{i, c}+\sum_{i=1}^N \sigma\left(z_{i, c}\right)+\epsilon}\right), \\
    \mathcal{L}_{\mathrm{CE}}=-\frac{1}{N} \sum_{i=1}^N \sum_{c=1}^C y_{i, c} \log p_{i, c}, \quad p_{i, c}=\frac{\exp \left(z_{i, c}\right)}{\sum_{j=1}^C \exp \left(z_{i, j}\right)}.
    \label{DiceCELoss}
\end{aligned}
\end{equation}
This loss function is a sum of both the Dice loss and the cross-entropy loss. In \eref{DiceCELoss}, $C$ denotes the total number of classes, $N$ represents the number of pixels in one image, $y_{i, c}$ is the one-hot encoded true label for pixel $i$ and channel $c$, $z_{i, c}$ is the output logit for pixel $i$ and channel $c$, $\sigma$ is the sigmoid function, and $\epsilon$ is a small constant added to prevent division by zero.

\begin{figure}[h!] 
    \centering
         \includegraphics[width=1.0\textwidth]{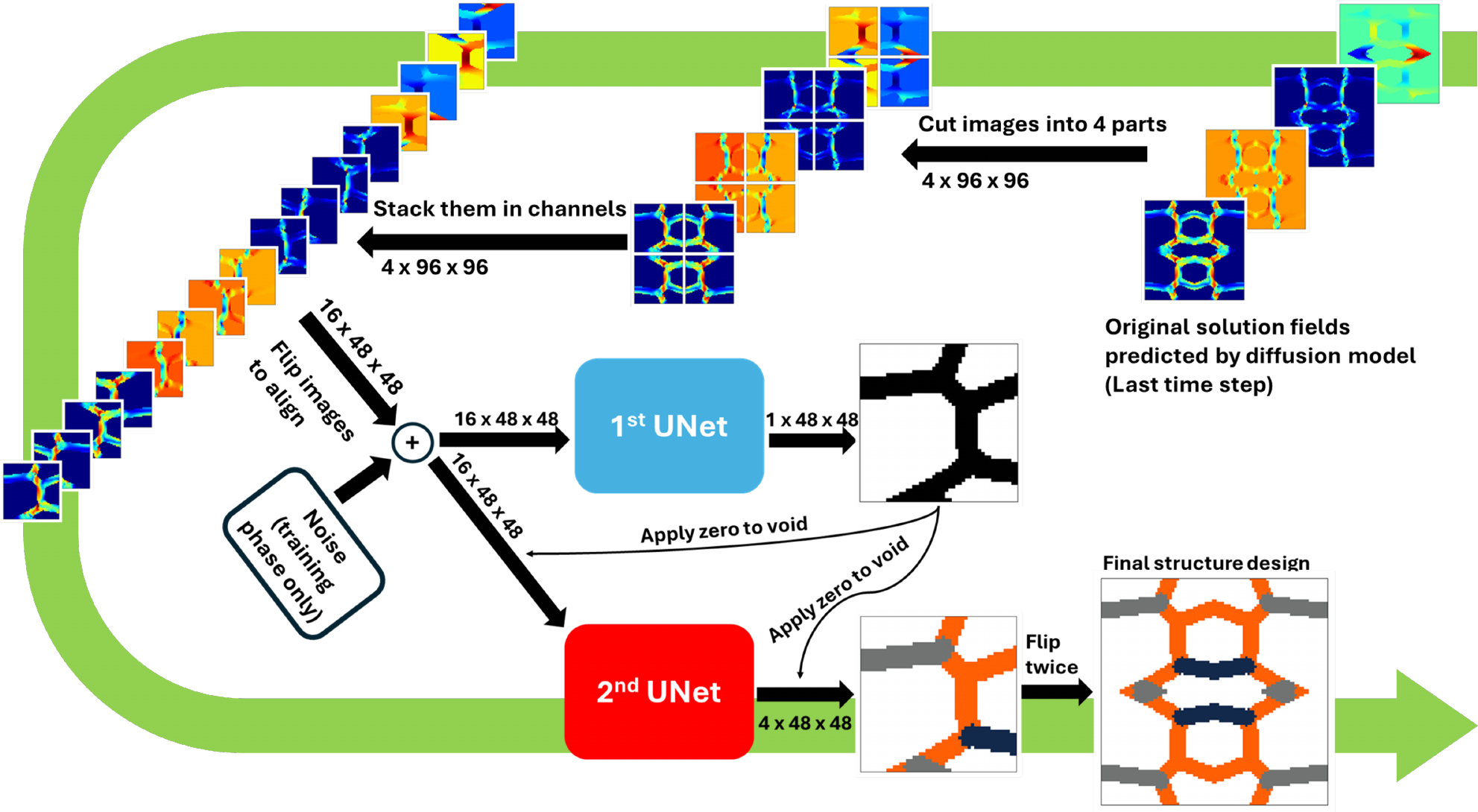}
    \caption{Two UNets after diffusion model as a structure identifier}
    \label{Two_UNets}
\end{figure}

The outputs from this network is a 48 $\times$ 48 image, each pixel segmented as one of the four possible classes (including void). A schematic of this UNet is the same with the first UNet figure as shown in \fref{UNet_structure}. The raw mechanical fields in the training dataset and predicted by the diffusion model are 96$\times$96 images, while both UNet models operate on 48$\times$48 images. This choice is deliberate since we would like the full designs (have shape 96$\times$96) to be periodic and symmetric, which can be achieved exactly by copying and flipping the one-quarter design (i.e., the 48$\times$48 image). As such, each input field are first sliced into 4 quarters and rotated to a consistent orientation, turning each 96$\times$96 input image into shape 48$\times$48$\times$4. The complete workflow for lattice structure generation given a user-defined stress-strain curve is shown in \fref{Two_UNets}. 

\section{Results and discussion}
\label{sec:results}
% As already stated, all simulations involving physics and finite element analysis to generate data were conducted using Abaqus/Standard \citep{Abaqus2022} on an AMD EPYC 7763 Milan CPU cores. Additionally, all deep learning training tasks were executed on the Nvidia A100 GPU nodes within the Delta cluster. 

\subsection{Diffusion outputs and their noisy nature}
\begin{figure}[h!] 
    \centering
         \includegraphics[width=0.7\textwidth]{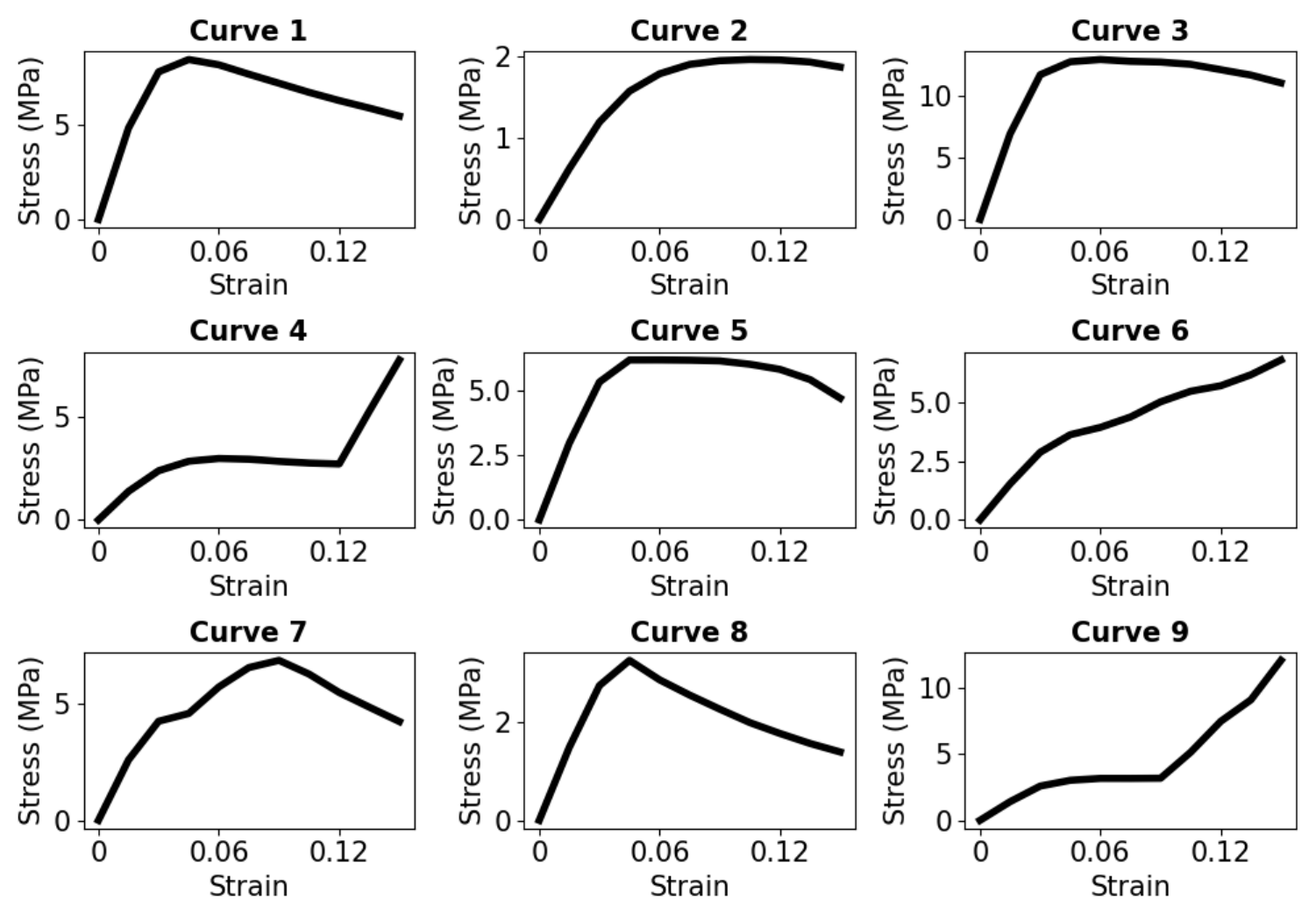}
    \caption{Nine target curves chosen for the diffusion model to mimic.}
    \label{Target_curves}
\end{figure}

To evaluate the diffusion model's performance across various mechanical behaviors, nine diverse stress-strain curves were selected from finite element simulation results. These curves, shown in \fref{Target_curves}, serve as target behaviors for the framework to reproduce similar designs.
Curves 1 and 5 exhibit conventional elasto-plastic behavior with distinct elastic and plastic regions, including strain hardening and softening. Curve 2 displays a significant plateau in the plastic region, indicative of perfect plasticity or a prolonged yielding phase. Curve 3 mirrors the behavior of Curve 2 but at a five-times higher magnitude, testing the framework's ability to scale behaviors. Curves 4 and 9 feature abrupt stress jumps, indicating densification of the structure at larger strains. Curve 6 demonstrates monotonic hardening throughout the strain range, a behavior seen in some composite materials or during certain thermomechanical processes. Curve 7 presents a complex response with two distinct slopes separated by a plateau, potentially representing a material with multiple deformation mechanisms or phase changes. Lastly, Curve 8 showcases a pronounced stress decrease after reaching a maximum, typical of materials exhibiting significant softening due to buckling or fracture. This diverse set of target curves encompasses a wide range of mechanical responses, challenging the framework to reproduce behaviors from conventional plasticity to more complex, non-linear phenomena. Such variety ensures a robust evaluation of the framework's capabilities in mimicking realistic material behaviors.

\begin{figure}[h!]
    \centering
    \begin{minipage}[b]{0.35\textwidth}
        \centering
        \includegraphics[width=\textwidth]{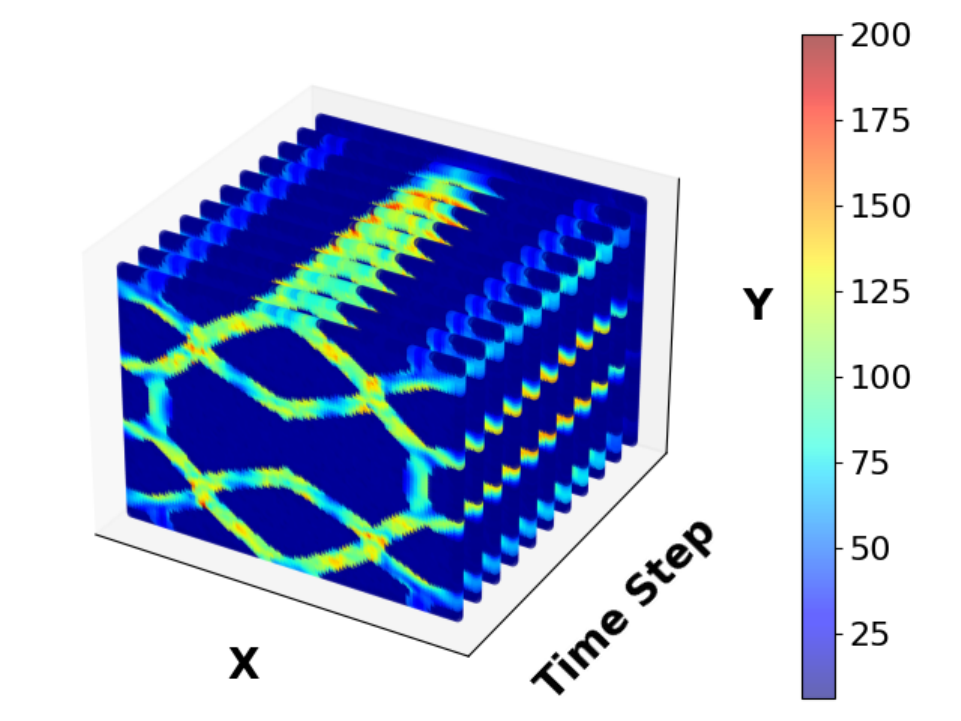}
        \textbf{(a)} $\sigma_{vM}$
    \end{minipage}
    %\hfill
    \begin{minipage}[b]{0.35\textwidth}
        \centering
        \includegraphics[width=\textwidth]{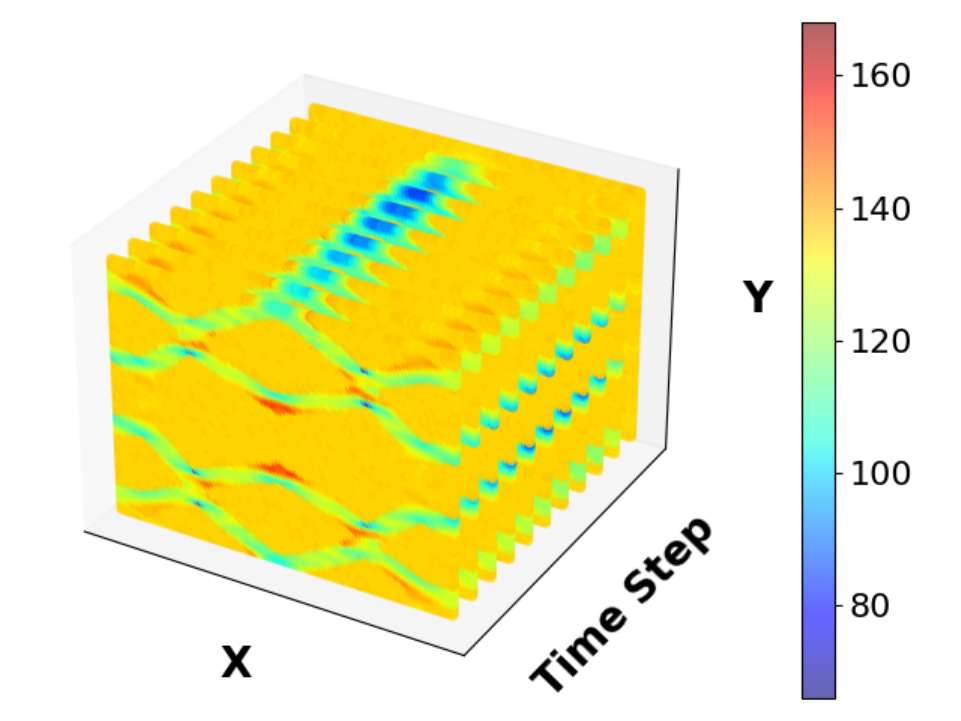}
        \textbf{(b)} $\sigma_{yy}$
    \end{minipage}
    
    \vskip\baselineskip
    \begin{minipage}[b]{0.35\textwidth}
        \centering
        \includegraphics[width=\textwidth]{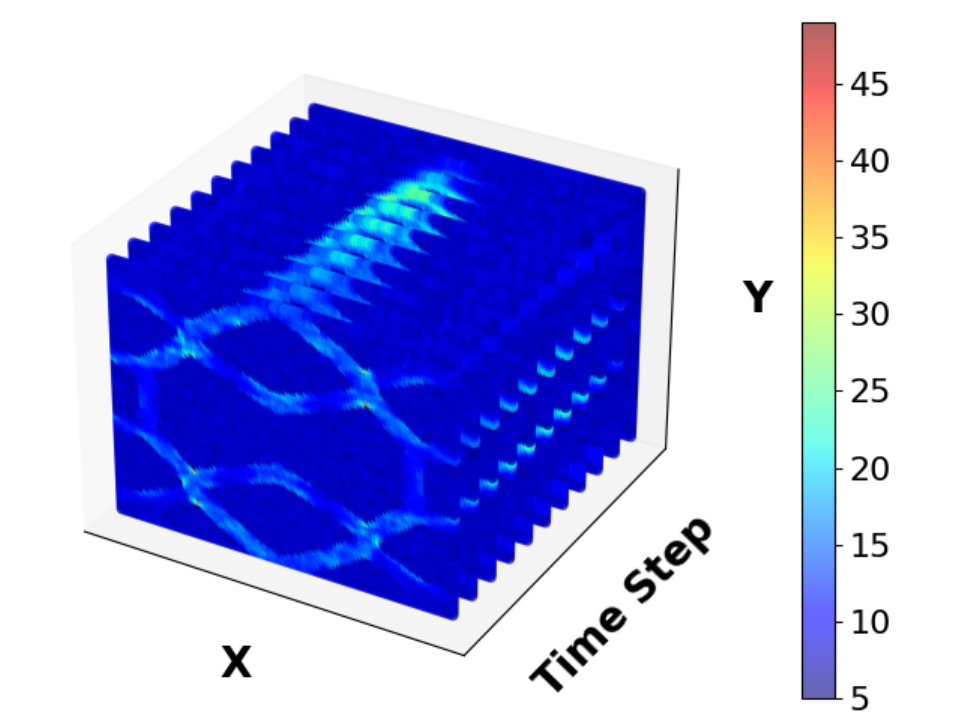}
        \textbf{(c)} $g_{se}$
    \end{minipage}
    %\hfill
    \begin{minipage}[b]{0.35\textwidth}
        \centering
        \includegraphics[width=\textwidth]{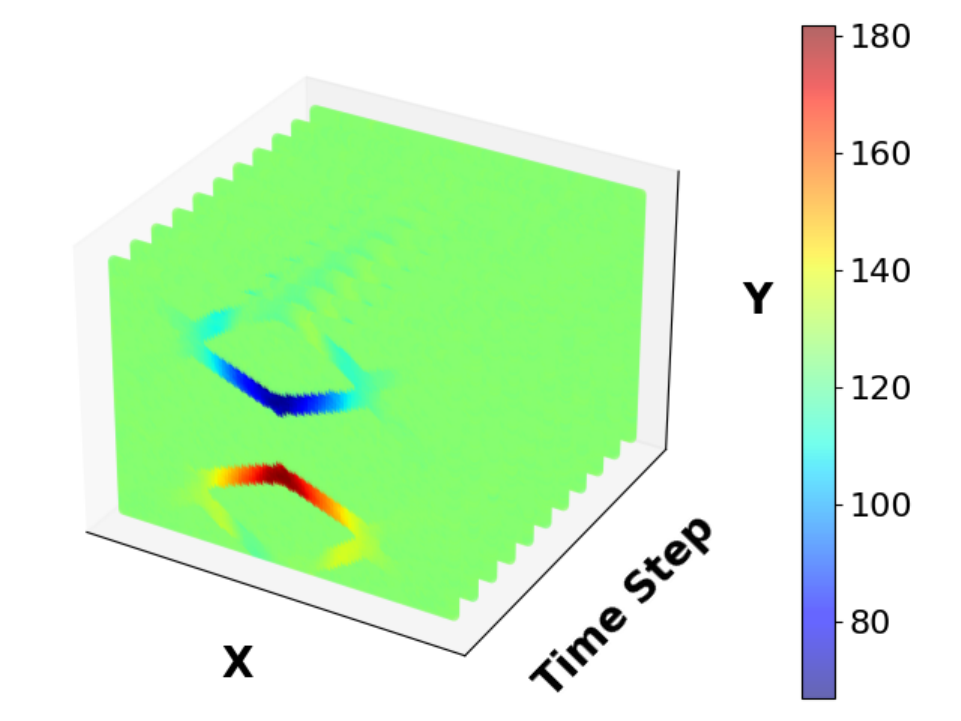}
        \textbf{(d)} $u_{x}$
    \end{minipage}

    \caption{Diffusion raw output for a single sample with 11 time steps in each of the 4 fields.}
    \label{fig:diffusion_raw_full_timestep}
\end{figure}

\begin{figure}[h!] 
    \centering
    \subfloat[Normalized raw field output from diffusion model, last time step]{
         \includegraphics[trim={0cm 0cm 0cm 0cm},clip,width=0.95\textwidth]{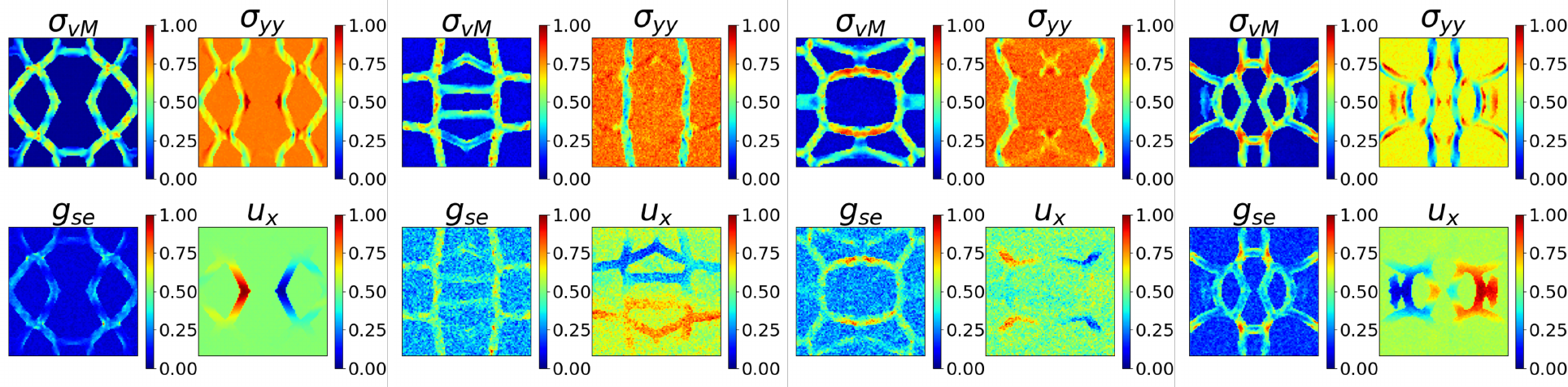}
         \label{Diffusion_noisy}
     }
    \\
    \subfloat[Normalized fields extracted from FEA, last time step]{
         \includegraphics[trim={0cm 0cm 0cm 0cm},clip,width=0.95\textwidth]{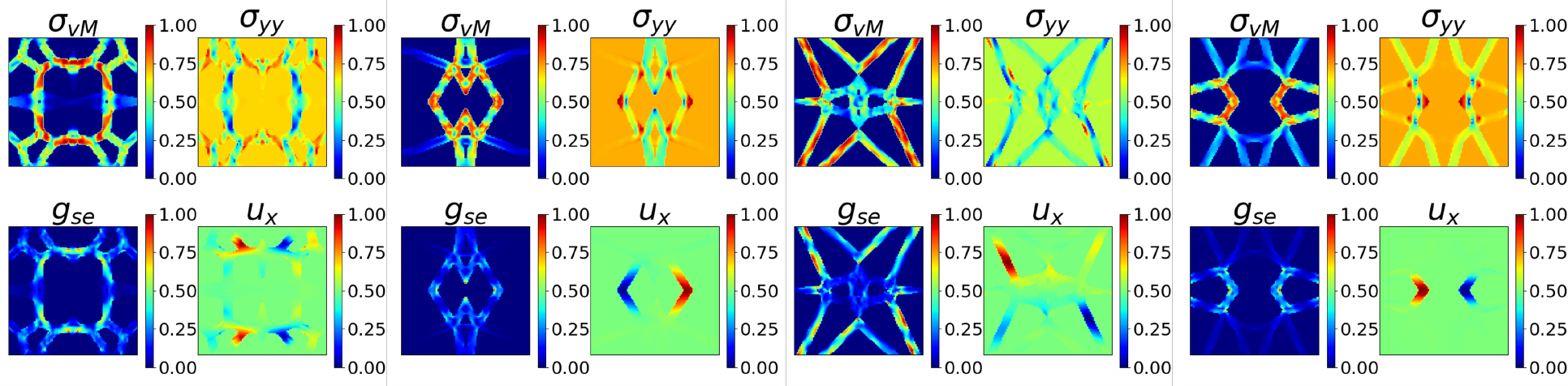}
         \label{Abaqus_clean}
     }
     
    \caption{Visual noise comparison of field values obtained from diffusion model and FEA. }
    \label{Noise_clean_comnpare}
\end{figure}

The diffusion model was trained over 180,000 iterations with 3314 simulations. The training process underwent 21 hours to complete, and the following sampling process took approximately 4 hours to generate 450 samples. \fref{fig:diffusion_raw_full_timestep} presents an illustrative example of the raw output generated by the diffusion model, using the target curves in \fref{Target_curves}. Consistent with its training paradigm, the model produces four distinct fields for each sample. A prominent characteristic of these outputs is the presence of Gaussian noise, a consequence of the model's inherently stochastic nature. While the iterative denoising process within the diffusion model aims to mitigate artifacts and eliminate noise, the raw output often retains residual noise. This noise can impede the structure identifier's ability to accurately discern the lattice design. For example, \fref{Noise_clean_comnpare} illustrates the contrast in noise levels between diffusion model outputs and FEA for the last time step. The figure presents two sets of normalized samples: \fref{Diffusion_noisy} shows outputs from the diffusion model, while \fref{Abaqus_clean} displays results from FEA. Visual inspection reveals that the fields generated by the diffusion model exhibit significantly more granularity compared to the smoother, cleaner outputs obtained from FEA. This disparity in image quality underscores the challenges posed by the noise in diffusion model outputs and highlights the need for strategies in countering the residual noise in the prediction for practical applications.

To characterize the noise level in the diffusion-generated data, several metrics can be employed to quantify noise within a single image. Typically, measuring noise in an image requires two versions: an original and a noisy version of the same image. However, in our work, there are no "original" versions for the diffusion-generated data, as these images are newly created. Therefore, a noise measure that can be computed using just one image is necessary. Discrete cosine transform (DCT) energy \citep{soon1998noisy} was utilized in this work to evaluate a noise in a single image. The DCT energy is obtained by transforming an image using the DCT and summing the absolute values of the transformed pixel values. Since this measure can represent the frequency content of an image, a noise-free image will result low DCT energy value due to its lower-frequency components. Conversely, a noisy image will result a high DCT energy value because noise typically spreads across higher frequencies.

\begin{figure}[h!] 
    \centering
    \subfloat[DCT energy of diffusion model output]{
         \includegraphics[trim={0cm 0cm 0cm 0cm},clip,width=0.47\textwidth]{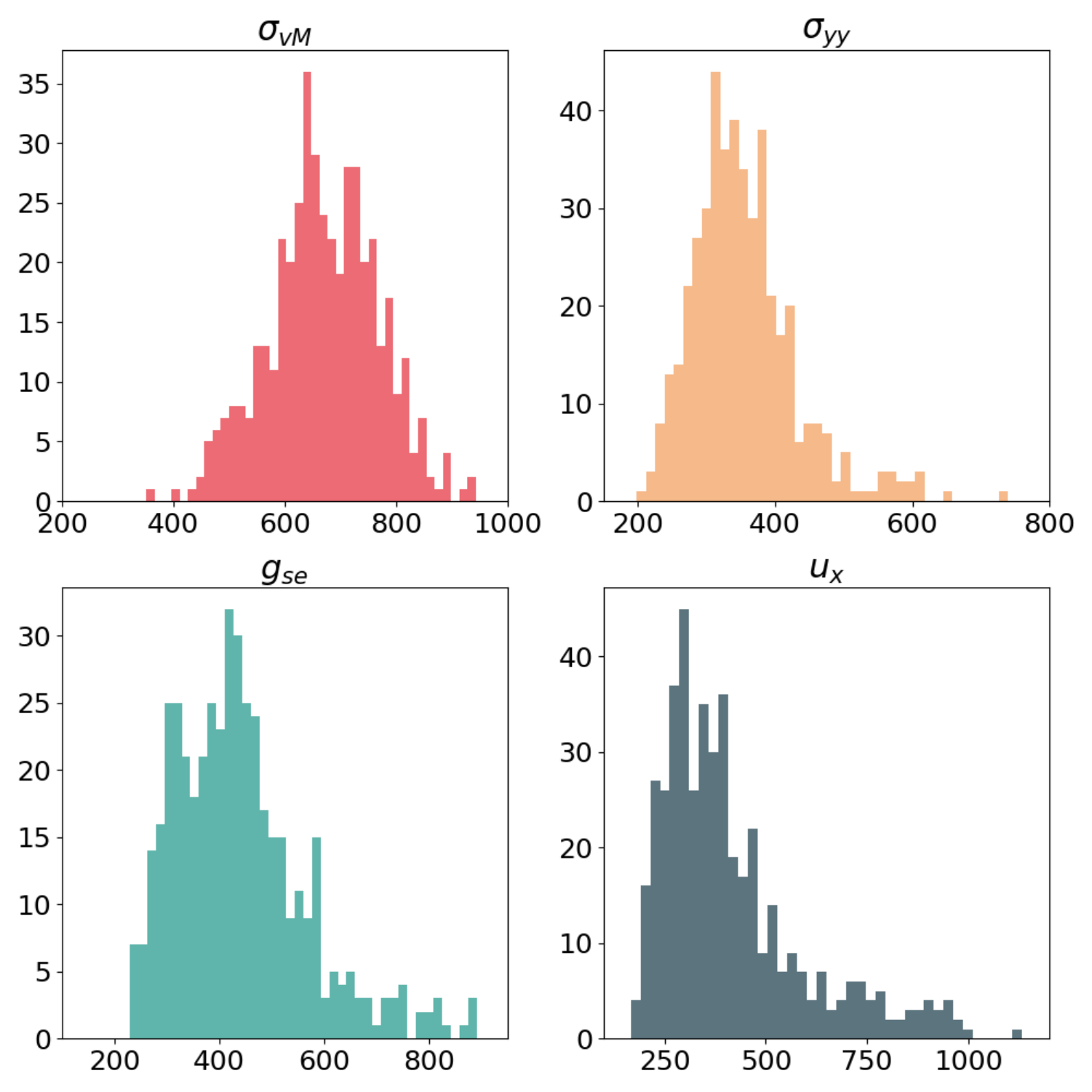}
         \label{DCT_Diffusion}
     }
    \hspace{0.1cm}
    \subfloat[DCT energy of FEA fields]{
         \includegraphics[trim={0cm 0cm 0cm 0cm},clip,width=0.47\textwidth]{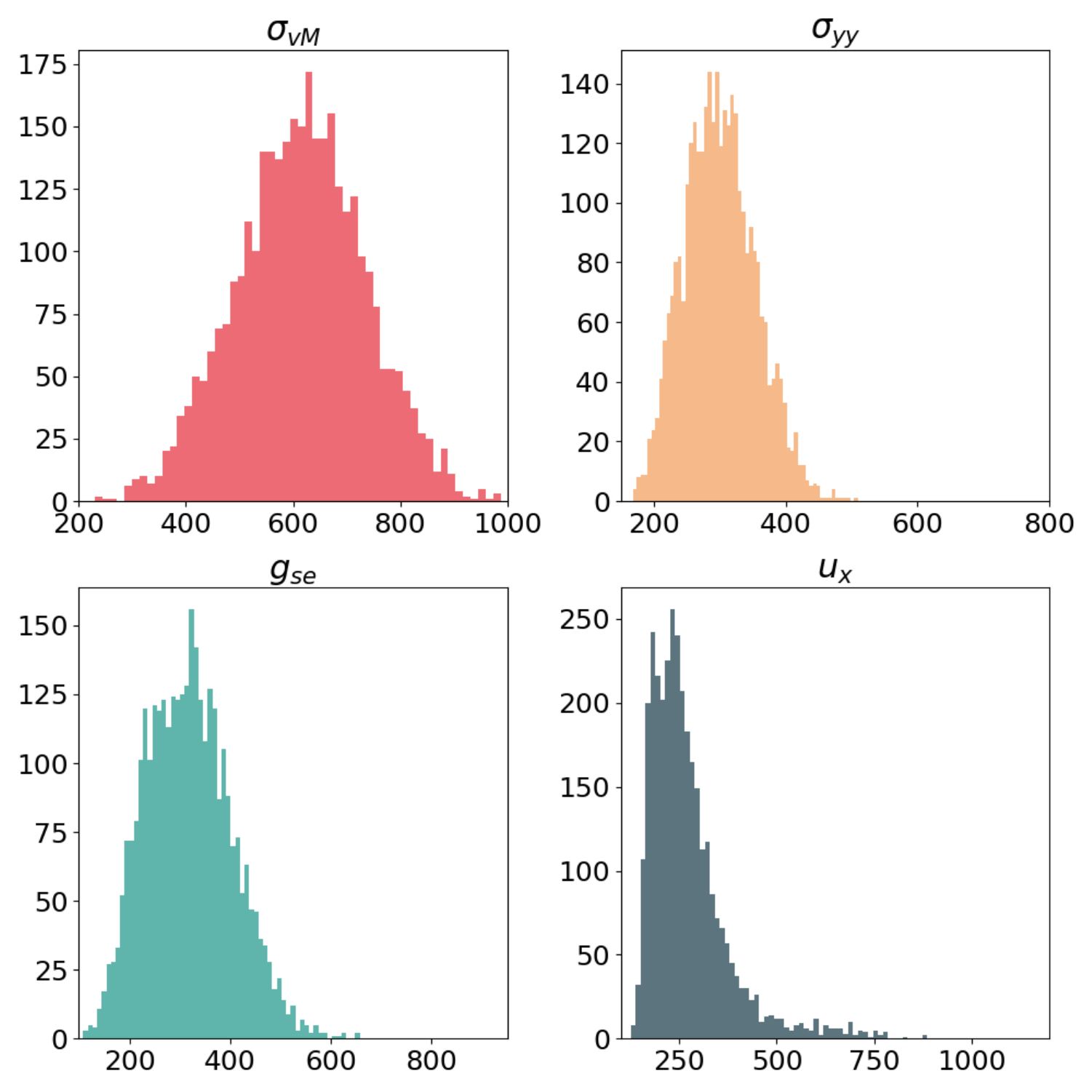}
         \label{DCT_Abaqus}
     }
     
    \caption{Histograms of DCT energy for each field property, last time step in the two datasets: (a) diffusion model outputs and (b) FEA. The DCT energy metrics were computed for individual field properties to compare the noise characteristics between the datasets.}
    \label{DCT_compare}
\end{figure}

\fref{DCT_compare} compares the DCT energy of the last time step between the diffusion model output and the FEA. It can be observed that the DCT energy values in the diffusion-generated data are generally higher across all four fields, indicating that these data contain more high-frequency components, which are likely attributed to noise. Additionally, while the DCT energy histogram for $\sigma_{vM}$ shows little difference between the two datasets, the $g_{se}$ and $u_x$ fields exhibit more significant differences. The corresponding statistics are summarized in \tref{tab:noise_stats}. In terms of mean values, the DCT energy for $\sigma_{vM}$ is 9\% higher in the diffusion model output compared to the FEA fields. For the $\sigma_{yy}$ field, it is 18\% higher, for the $g_{se}$ field, it is 38\% higher, and for the $u_x$ field, it is 52\% higher in the diffusion outputs. 
\begin{table}[h!]
\centering
\caption{DCT Energy Statistics for Both Datasets: Diffusion Generated and FEA, last time step}
\begin{tabularx}{\textwidth}{ 
  >{\centering\arraybackslash}X 
  >{\centering\arraybackslash}X 
}
\begin{tabular}{c|c|c}
\multicolumn{3}{c}{\textbf{Diffusion Generated}} \\ \hline
\textbf{Channel} & \textbf{Mean} & \textbf{Maximum} \\ \hline
$\sigma_{vM}$        & 671     & 943     \\ 
$\sigma_{yy}$        & 354     & 739     \\ 
$g_{se}$             & 440     & 891    \\ 
$u_x$                & 415     & 1130    \\ 
\end{tabular}
&
\begin{tabular}{c|c|c}
\multicolumn{3}{c}{\textbf{FEA}} \\ \hline
\textbf{Channel} & \textbf{Mean} & \textbf{Maximum} \\ \hline
$\sigma_{vM}$       & 615     & 1070    \\ 
$\sigma_{yy}$       & 300     & 510    \\ 
$g_{se}$            & 319     & 658    \\ 
$u_x$               & 272     & 887      \\ 
\end{tabular}

\end{tabularx}
\label{tab:noise_stats}
\end{table}

The residual noise within the diffusion output manifests in various ways, influenced by factors such as the lattice structure's complexity and the intrinsic characteristics of the simulated mechanical fields. Traditional denoising algorithms, like Gaussian filters or Block-Matching and 3D Filtering (BM3D), offer a potential solution for refining noisy raw outputs. These methods are computationally efficient and straightforward to implement. However, our approach leverages separate NNs in conjunction with the diffusion model to refine and identify the multi-material design. To implement this strategy, Gaussian noise was artificially introduced to the FEA training data (training dataset of diffusion model). The NNs described in \sref{sec:Two_UNets} were subsequently trained on this noise-enhanced data. This approach allows the networks to effectively handle the noisy raw outputs of the diffusion model during inference, as they have been conditioned on similarly noisy data during training. It is crucial that the noise characteristics added to the FEA training data comprehensively represent those present in the diffusion outputs. We selected Gaussian noise to align with the noise type used in the diffusion model. The intensity of the noise introduced to the training data was calibrated to match the highest noise levels observed in the diffusion outputs, as measured by their DCT energy. Specifically, the Gaussian noise was added to the FEA simulation data such that the maximum DCT energy values exceed at least with 25.0\% margin to those of the diffusion-generated last time step outputs for each field separately. For example, the DCT energy for $\sigma_{vM}$ in the noise-enhanced FEA training data was set at 125\% of the maximum value found in the diffusion-generated $\sigma_{vM}$ fields. Similarly, the fields $\sigma_{yy}$, $g_{se}$, and $u_x$ in the FEA training data were also augmented with Gaussian noise to reach 125\% of the respective maximum DCT energy levels of the diffusion-generated fields. This approach ensures that the structure identifier was trained on even noisier data than it would encounter during inference, preparing it to handle diffusion-generated fields effectively in the inference phase.

\tref{tab:noise_stats_second} summarizes the DCT energy statistics of the noise-enhanced FEA training data. The last column in this table (i.e., Max DCT Energy Ratio Noise-enhanced / Diffusion), provides a comparison between the noise-enhanced FEA simulations and the raw diffusion outputs. This ratio, consistently exceeding 100\% for all channels, demonstrates that our noise-enhanced training dataset indeed incorporates noisier data than the raw diffusion outputs. Specifically, the maximum DCT energy ratios range from 125\% to 127\% across all channels. This indicates that we have successfully introduced an additional 25.0\% noise (or even more) margin in our training data compared to the expected noise levels in the diffusion outputs. The step ensures that the NNs are trained on data with a higher noise level, improving their robustness and effectiveness when handling the noisy outputs of the diffusion model during the inference phase.
\begin{table}[h!]
\centering
\caption{DCT Energy Statistics for Noise added FEA simulation, last time step}
\begin{tabularx}{\textwidth}{ 
  >{\centering\arraybackslash}X 
  >{\centering\arraybackslash}X 
}
\begin{tabular}{c|c|c|c}
\multicolumn{4}{c}{\textbf{Noise added FEA Simulations}} \\ \hline
\textbf{Channel} & \textbf{Mean} & \textbf{Maximum} & \textbf{Max DCT Energy Ratio } \\ 
 & & & (Noise-enhanced FEA / Diffusion) \\ \hline
$\sigma_{vM}$     & 807      & 1190     & 127\%   \\ 
$\sigma_{yy}$     & 802      & 934      & 126\%   \\ 
$g_{se}$          & 856      & 1120     & 125\%   \\ 
$u_x$             & 1160     & 1420     & 125\%   \\ 
\end{tabular}

\end{tabularx}
\label{tab:noise_stats_second}
\end{table}

\begin{figure}[h!] 
    \centering
         \includegraphics[width=0.95\textwidth]{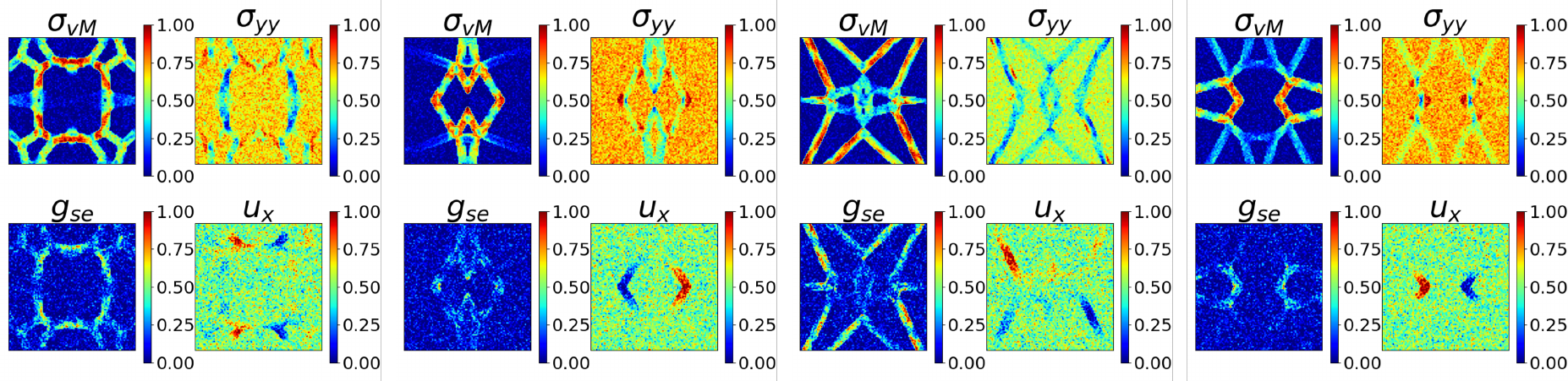}
    \caption{Four samples in the noise-enhanced FEA training data. Each sample contain 4 fields: $\sigma_{vM}$, $\sigma_{yy}$, $g_{se}$, and $u_x$. For comparison, same samples with \fref{Abaqus_clean} were taken, but with added noise.}
    \label{NAA_images}
\end{figure}

\fref{NAA_images} illustrates four samples in the noise-enhanced FEA training data. Visually, they resemble \fref{DCT_Diffusion} more closely than \fref{DCT_Abaqus}, which aligns with our intention to simulate the noise characteristics of the diffusion model outputs. The added Gaussian noise introduces a granular texture to the images, mimicking the stochastic nature of diffusion-generated samples. This visual similarity is crucial, as the structure identifier consists of convolutional layers and learn directly from images in the training data. The figure also indicates that our noise-enhancement strategy successfully bridges the gap between the clean FEA and the noisier diffusion outputs.

\subsection{Performance of the structure identifier}
The structure identifier (i.e., two UNets) were trained and evaluated using an 90-10 split of the dataset into training and test sets. From the 3314 simulations, 332 samples were randomly selected to form the test dataset. The training process consisted up to 500 epochs with stopping early if the test accuracy reaches 99.9\%. NN training was executed on an Nvidia A100 GPU. The first UNet took 47 minutes with 53 epochs resulting in an average iteration time of 52.8 seconds, and the second UNet took 3 hours with 178 epochs, having 62.4 seconds on average per iteration.

To evaluate the accuracy of the trained NNs, the following metrics were employed:
\begin{equation}
\begin{aligned}
    {\text{Accuracy}(\%)} = \frac{\sum_{p=1}^M \sum_{q=1}^M \mathds{1}\left(\text{Y}_{\text {true }, p, q}=\text{Y}_{\text {pred }, p, q}\right)}{M^2} \times 100 (\%), \\
    \text{Weighted DSC}(\%) = (\sum_{j=1}^{C} w_j \cdot \text{DSC}_j) \times 100(\%), \\
    \text{DSC}_j = 
    \begin{cases} 
    \frac{2 |\text{Y}_\text{true}^j \cap \text{Y}_\text{pred}^j|}{|\text{Y}_\text{true}^j| + |\text{Y}_\text{pred}^j|} & \text{if } |\text{Y}_\text{true}^j| + |\text{Y}_\text{pred}^j| > 0 \\
    1 & \text{if } |\text{Y}_\text{true}^j| + |\text{Y}_\text{pred}^j| = 0
    \end{cases}, \quad w_j = \frac{|\text{Y}_\text{true}^j|}{\sum_{k=1}^{C} |\text{Y}_\text{true}^k|}
    \label{model_performance_metric}
\end{aligned}
\end{equation}
where $M \times M$ is the size of each 2D design image, $\mathds{1}$ is the indicator function which returns 1 if the condition is true and 0 otherwise. $\text{Y}_{\text {true }}, \text{Y}_{\text {pred }}$ are true and predicted images respectively. The $(p, q)$ subscript in the first equation denotes $(p, q)$-pixel value in $\text{Y}_{\text {true }} $ and $ \text{Y}_{\text {pred }}$. In the second equation,  $C$ is the number of classes, $w_j$ is the weight assigned to class $j$. This weight is calculated as the ratio of the number of pixels belonging to class $j$ in the true segmentation map $\text{y}_{ture}$ to the total number of pixels in all classes. Similarly, $\text{DSC}_{j}$ represents the Dice Similarity Coefficient (DSC) for class $j$, obtained from the third row of \eref{model_performance_metric}. The numerator represents two times the intersection of the true and predicted segmentation for class $j$ (denoted as $\text{Y}_{\text{true}}^j$ and $\text{Y}_{\text{true}}^j$), and the denominator is the total number of pixels labeled as class $j$ in both the ground truth and the predicted segmentation. The accuracy and weighted DSC metrics offer a thorough assessment of the model's performance in predicting both material mask and multi-material design, by a quantitative comparison between the NN's predictions and the ground truth data from the design itself. The DSC equation was modified to encompass multi-class case, calculating weighted DSC values for each class and averaging them.

With the early-stopping mechanism employed, both UNet models did not require the full 500 epochs to achieve 99.9\% accuracy. The first UNet, tasked with binary segmentation, reached the target accuracy of 99.9\% after 53 epochs. The second UNet, responsible for multi-class segmentation, achieved the same target accuracy after 178 epochs. The test dataset results demonstrate that the structure identification architecture attained a consistently high accuracy in predicting designs based on their mechanical fields, effectively extracting designs from the noisy raw outputs of the diffusion model. The following \tref{tab:structure_identifier_per} provides a detailed overview of the performance metrics, as defined in \eref{model_performance_metric}, for the two UNets within the structure identifier model across various percentiles, including the best case, 25$^{th}$ percentile, 50$^{th}$ percentile (median), 75$^{th}$ percentile, and the worst-case scenario. The results demonstrate that the structure identifier exhibits robust performance on the test dataset, which notably contains samples with noise levels exceeding those anticipated in the inference data.

\begin{table}[h!]
\centering
\caption{Structure identifier performance metrics}
\begin{tabularx}{\textwidth}{ 
  >{\centering\arraybackslash}X 
  >{\centering\arraybackslash}X 
}
\begin{tabular}{c|c|c|c|c|c}
\multicolumn{6}{c}{} \\ \hline
\textbf{UNet-1: binary} & \textbf{Best} & \textbf{25$^{th}$} & \textbf{50$^{th}$} & \textbf{75$^{th}$} & \textbf{Worst} \\ \hline
\textbf{Accuracy}  & 100\% & 100\% & 100\% & 99.9\% & 98.6\%    \\ \hline
\textbf{Weighted DSC}       & 100\% & 100\% & 100\% & 99.9\% & 99.5\%
\end{tabular}
\begin{tabular}{c|c|c|c|c|c}
\multicolumn{6}{c}{} \\ \hline
\textbf{UNet-2: multi-class} & \textbf{Best} & \textbf{25$^{th}$} & \textbf{50$^{th}$} & \textbf{75$^{th}$} & \textbf{Worst} \\ \hline
\textbf{Accuracy} & 100\% & 100\% & 99.9\% & 99.9\% & 99.5\%    \\ \hline
\textbf{Weighted DSC}      & 100\% & 99.9\% & 99.9\% & 99.9\% & 99.5\%
\end{tabular}
\end{tabularx}
\label{tab:structure_identifier_per}
\end{table}

\begin{figure}[h!] 
    \centering
         \includegraphics[width=1.0\textwidth]{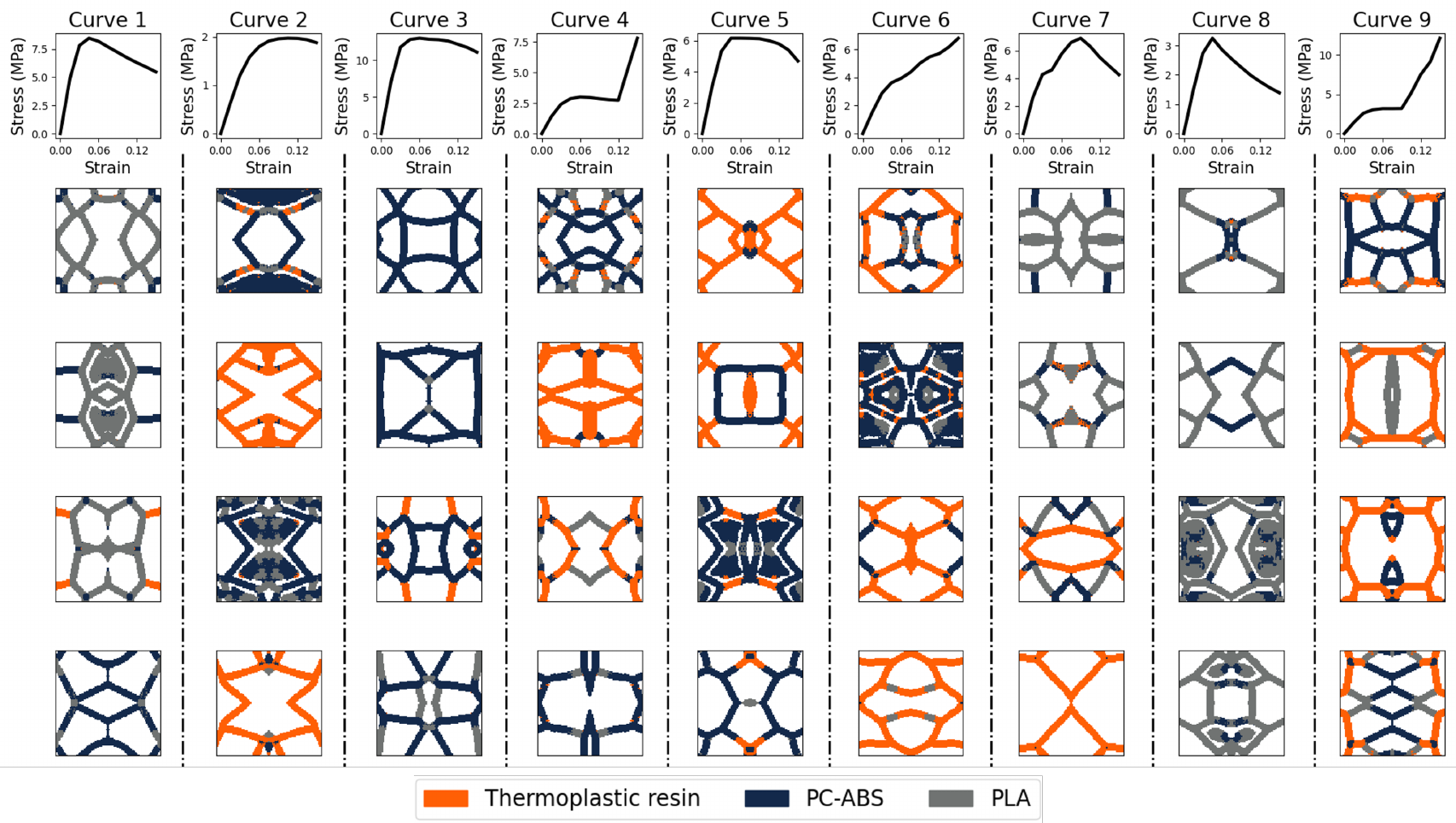}
    \caption{Designs created by NNs corresponding to the target curve at the first row. Five were chosen randomly out of 50 designs generated for each target curve.}
    \label{Diffusion_designs_configured}
\end{figure}

Using the trained model, an inference process was conducted on the raw output from the diffusion model to obtain the 2D multi-material designs. Given the stochastic nature of input noise image to the diffusion model, 50 designs were generated from each target curve, leading to a total of 450 designs. \fref{Diffusion_designs_configured} shows 5 randomly selected multi-material designs for each target curve, showing the diversity of the designs generated from the current framework.

\subsection{FEA validation of designs produced by NNs}
Ensuring that these designs are manufacturable and closely resemble the initial Voronoi designs used during the training data generation phase is crucial. To evaluate the manufacturability and plausibility of designs generated by the diffusion and structure identifier framework, we conducted the finite element analysis identical to that described in \sref{sec:training_data_generation}. The FEA results, particularly the stress-strain curves, were compared against the original target curves used to guide the diffusion model. For each target curve, 50 designs were generated, resulting in 50 corresponding stress-strain curves from FEA. The similarity between these generated curves and the target curve was assessed to gauge the effectiveness of the generative AI model in inherently mimicking the desired mechanical properties.
To quantify the closeness between the generated design curves and their corresponding target curves, we employed the metrics detailed in \eref{Curve_metric}. Let $y_{true}$ and $y_{pred}$ represent the target curve and the curves from generated designs, respectively, and $S$ denote the number of points in both curves. As defined in \eref{Curve_metric}, we utilized the Relative Root Mean Squared Error (RRMSE) and the Relative Mean Absolute Error (RMAE) to provide a comprehensive analysis of the curve similarities. This approach allowed us to objectively assess the performance of our generative model in producing designs that closely match the desired mechanical properties specified by the target curves.

\begin{equation}
\begin{aligned}
    {\text{RRMSE}(\%)} = \frac{||y_{true} - y_{pred}||_{2}}{||y_{true}||_{2}} \times 100 (\%) \\
    {\text{RMAE}(\%)} =\frac{\sum_{i=1}^{S} |y_{true, i} - y_{pred, i}|}{\sum_{i=1}^{S} |y_{true, i}|} \times 100 (\%)
    \label{Curve_metric}
\end{aligned}
\end{equation}

The RRMSE results for all nine target curves are tabulated in \tref{tab:RRMSE_curves}. These results demonstrate that the stress-strain curves derived from the NN-generated designs generally exhibit strong alignment with the target curves. Analysis of the RRMSE for the best-matching designs reveals an average difference of 7.41\% and a median difference of 5.20\% between the target and generated design curves over all nine target curves. Notably, designs associated with Curve 3 display RRMSE values ranging from 3.77\% to 4.06\%, indicating a particularly high level of accuracy for this mechanical behavior. However, some curves, such as Curve 9, show higher variability in their generations, with RRMSE values reaching up to 14.2\% for the third-best match. These findings suggest that while the generative model performs well in most cases, there is some variation in its accuracy across different mechanical situations. This variability may provide insights into the model's strengths and limitations in replicating specific mechanical properties.

\begin{table}[h!]
\centering
\caption{RRMSE between the target curve and curves from generated designs. The first three best matches in RRMSE metric are shown.}
\footnotesize
\begin{tabularx}{\textwidth}{ 
  >{\centering\arraybackslash}X 
  >{\centering\arraybackslash}X 
}
\begin{tabular}{c|c|c|c|c|c|c|c|c|c}
\multicolumn{10}{c}{} \\ \hline
 & \textbf{Curve 1} & \textbf{Curve 2} & \textbf{Curve 3} & \textbf{Curve 4} & \textbf{Curve 5} & \textbf{Curve 6} & \textbf{Curve 7} & \textbf{Curve 8} & \textbf{Curve 9}\\ \hline
\textbf{Best}      & 5.20\% & 4.60\% & 3.77\% & 11.8\% & 4.92\% & 9.36\%  & 8.86\%  & 5.16\%  & 13.1\% \\ 
\textbf{$2^{nd}$}  & 6.58\% & 4.77\% & 4.00\% & 20.0\% & 6.60\% & 10.4\% & 10.9\% & 9.53\%  & 13.5\% \\ 
\textbf{$3^{rd}$}  & 7.25\% & 5.99\% & 4.06\% & 21.0\% & 7.10\% & 11.6\% & 11.6\% & 12.1\% & 14.2\% \\ 
\end{tabular}
\end{tabularx}
\label{tab:RRMSE_curves}
\end{table}

\begin{figure}[h!] 
    \centering
    \includegraphics[width=0.95\textwidth]{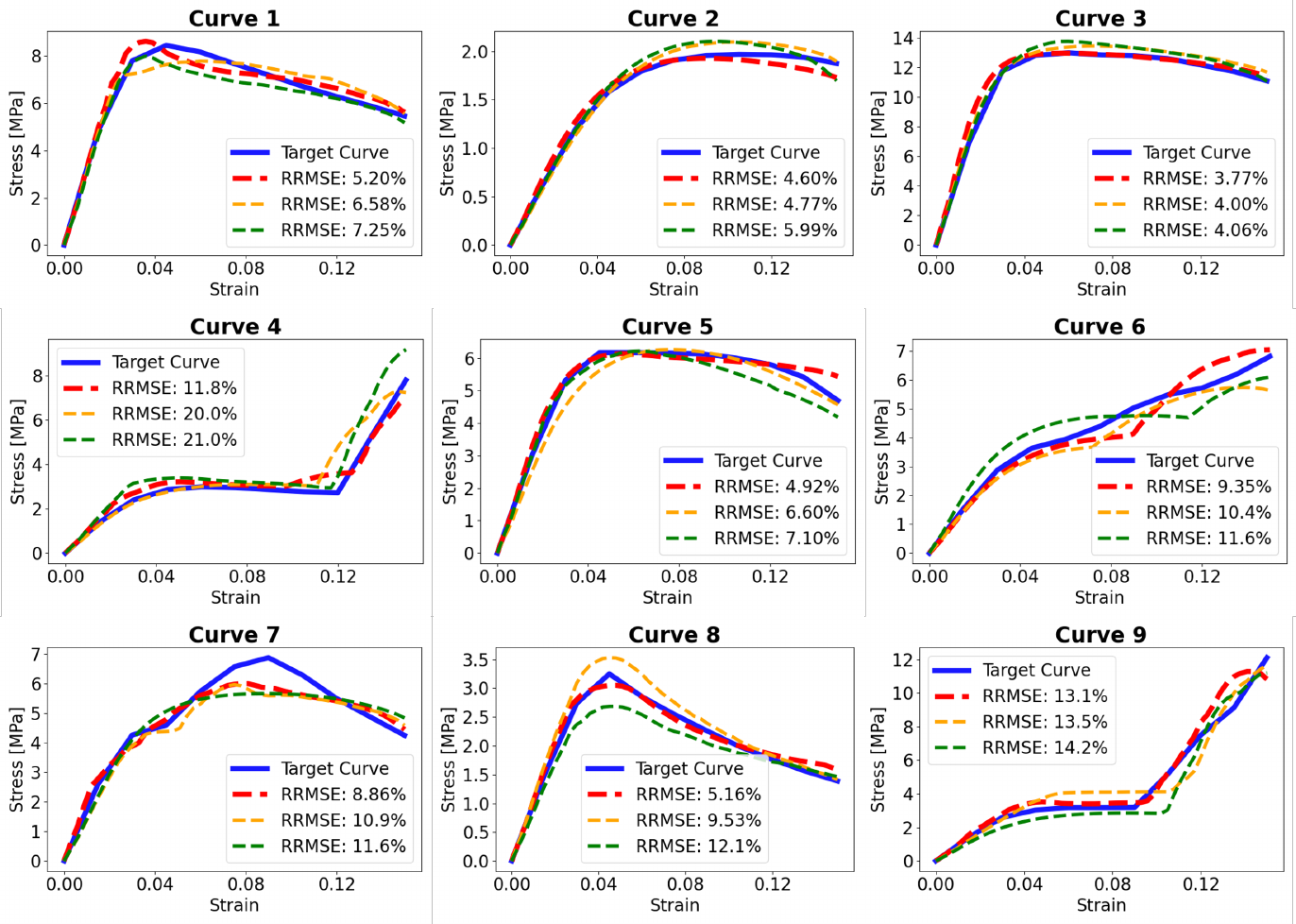}
    \caption{Comparison of target curves (solid blue) with curves from the three best-matching generated designs for each target curve, as measured by RRMSE. The dotted lines represent the generated designs, with red denoting the best match, orange the second-best match, and green the third-best match across all cases.}
    \label{fig:target_vs_generated_curves_RRMSE}
\end{figure}

\fref{fig:target_vs_generated_curves_RRMSE} provides a visual comparison between the target curves and the curves generated by our framework. For the majority of cases, the stress-strain curves produced by the generated designs closely align with the target curves, demonstrating the general effectiveness of our framework in capturing diverse mechanical behaviors. However, Curves 7 and 9 exhibit more noticeable deviations, indicating that certain mechanical behaviors pose greater challenges for our model to reproduce accurately.
Of particular interest are the sudden jumps observed in Curves 3 and 8. While the generated curves approximate this behavior, they struggle to capture the precise moment and magnitude of these abrupt changes. The visual results corroborate the RRMSE values presented in \tref{tab:RRMSE_curves}, providing a comprehensive view of the model's performance across various target mechanical behaviors. The analysis underscores both the strengths of our framework in reproducing a wide range of mechanical properties and the specific challenges that remain, particularly capturing the precise moment of the abrupt change in the mechanical behavior.

Following the RRMSE analysis, we employed the RMAE metric, as defined in \eref{Curve_metric}, to further evaluate the similarity between the target curves and those derived from the generated designs. The RMAE values for the best match, second-best match, and third-best match were calculated and tabulated for comprehensive analysis. The results are summarized in \tref{tab:NMAE_curves}. This metric complements the RRMSE analysis by offering a measure that is less sensitive to outliers, thereby providing a more robust assessment of the overall curve similarity. Additionally, RMAE offers a more intuitive interpretation of the average relative error between the target and generated curves.
This dual-metric approach allows to capture different aspects of curve similarity, enhancing the reliability of our performance assessment.

\begin{table}[h!]
\centering
\caption{RMAE between the target curve and curves from generated designs. The first three best matches in RMAE metric are shown.}
\footnotesize
\begin{tabularx}{\textwidth}{ 
  >{\centering\arraybackslash}X 
  >{\centering\arraybackslash}X 
}
\begin{tabular}{c|c|c|c|c|c|c|c|c|c}
\multicolumn{10}{c}{} \\ \hline
 & \textbf{Curve 1} & \textbf{Curve 2} & \textbf{Curve 3} & \textbf{Curve 4} & \textbf{Curve 5} & \textbf{Curve 6} & \textbf{Curve 7} & \textbf{Curve 8} & \textbf{Curve 9}\\ \hline
\textbf{Best}        & 4.60\% & 4.03\% & 2.17\% & 10.7\% & 3.44\% & 8.07\% & 7.20\% & 4.41\% & 11.0\% \\ 
\textbf{$2^{nd}$}    & 5.63\% & 4.33\% & 3.57\% & 13.7\% & 5.00\% & 8.95\% & 9.36\% & 8.23\% & 13.5\% \\ 
\textbf{$3^{rd}$}    & 5.80\% & 5.04\% & 3.82\% & 19.1\% & 5.73\% & 10.2\% & 9.91\% & 10.2\% & 14.4\% \\ 
\end{tabular}
\end{tabularx}
\label{tab:NMAE_curves}
\end{table}

The RMAE results presented in \tref{tab:NMAE_curves} exhibit a trend similar to the RRMSE analysis. For the best-matching designs across all nine target curves, the RMAE analysis reveals an average difference of 6.18\% and a median difference of 4.60\% between the target and generated-design curves. 

\begin{figure}[h!] 
    \centering
    \includegraphics[width=0.95\textwidth]{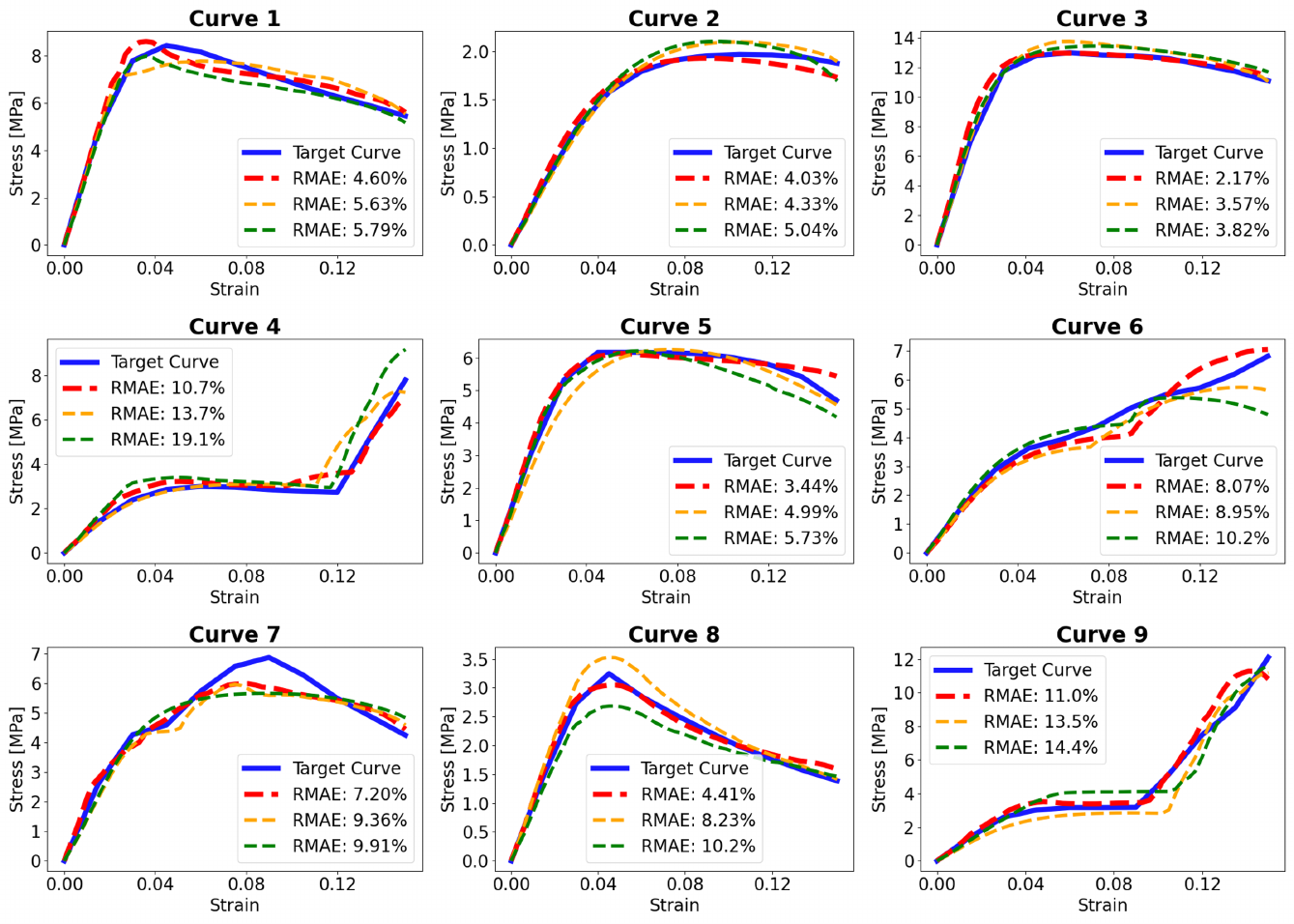}
    \caption{Comparison of target curves (solid blue) with curves from the three best-matching generated designs for each target curve, as measured by RMAE. The dotted lines represent the generated designs, with red denoting the best match, orange the second-best match, and green the third-best match across all cases.}
    \label{fig:target_vs_generated_curves_RMAE}
\end{figure}

\begin{figure}[h!] 
    \centering
    \includegraphics[width=0.85\textwidth]{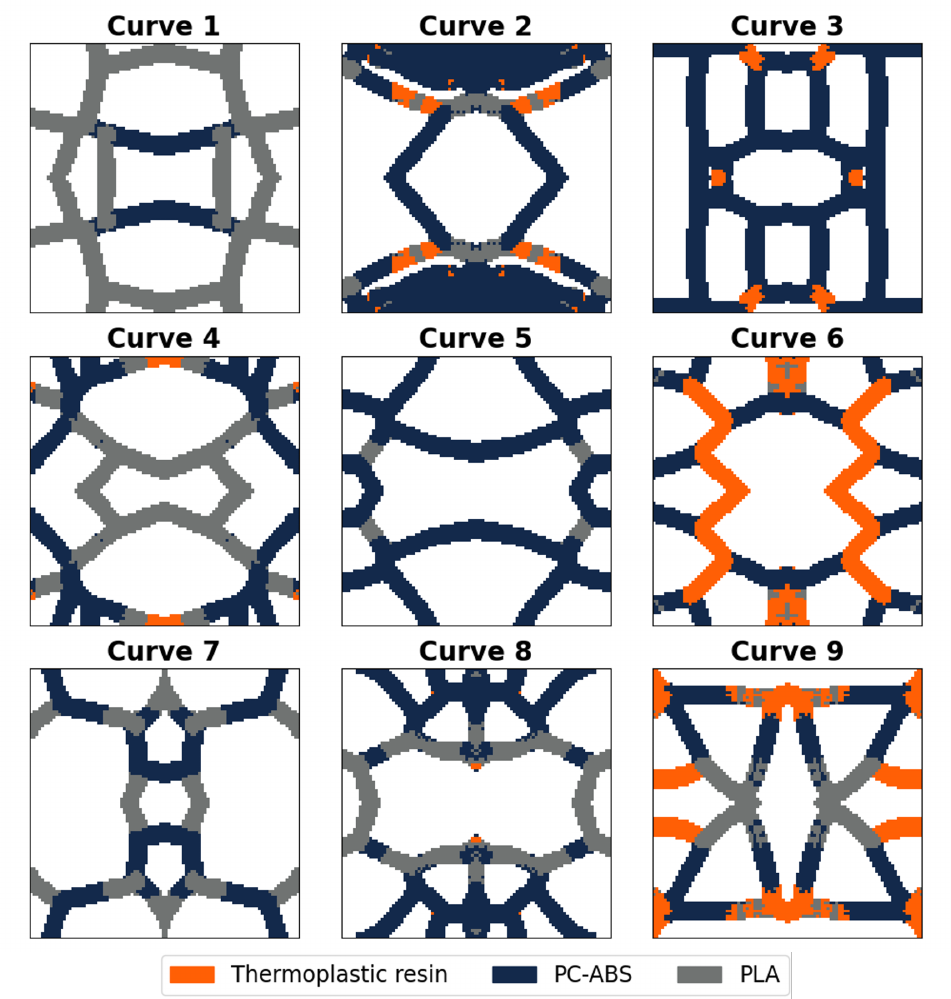}
    \caption{Designs that produced closest stress strain curve to the target curve measured by RRMSE and RMAE}
    \label{fig:Best_designs_RRMSE}
\end{figure}

Although RMAE generally yielded slightly lower values than RRMSE due to its reduced sensitivity to large individual errors, the relative performance across different target curves remained consistent between the two metrics. For instance, curves 2, 3, and 5 demonstrated low RMAE values, corresponding to their low RRMSE values. This consistency underscores the alignment between the two metrics and suggests that the model is particularly adept at reproducing these specific mechanical behaviors. Conversely, curves with sudden changes, such as 4 and 9, exhibited relatively higher RMAE values, aligning with their RRMSE results. Despite these challenges, all cases maintained RMAE values below 11\% for the best match, indicating reliable reproduction across various mechanical behaviors. \fref{fig:target_vs_generated_curves_RMAE} displays plots similar to those in \fref{fig:target_vs_generated_curves_RRMSE}, but using RMAE as the metric. Notably, the best-match designs identified by RRMSE and RMAE were identical for all nine target curves, further validating the consistency of our model's performance across different error metrics. Consequently, the best-match designs for RMAE are also represented in \fref{fig:Best_designs_RRMSE}.

Our framework demonstrates a remarkable capability to generate diverse and non-intuitive multi-material lattice designs, as illustrated in \fref{fig:Best_designs_RRMSE}. Of particular interest is the best design for Curve 2, which presents a unique hourglass-like structure with intersecting curved elements. This design deviates from conventional lattice patterns, featuring a central diamond shape bordered by asymmetric curved members. Such non-intuitive structures highlight the AI model's ability to explore complex design spaces beyond human preconceptions, potentially uncovering novel structures with desired mechanical properties. This competence in generating unexpected yet potentially high-performing designs accentuates the value of AI-driven approaches. These findings demonstrate the NN's effectiveness in generating innovative designs that closely match target stress-strain curves, highlighting its potential for applications in topology optimization and additive manufacturing. The consistency of the generated designs, with an average of 7.41\% RRMSE and 6.18\% RMAE, underscores the robustness of this approach. However, cases like Curve 7, which show higher RRMSE, indicate areas for further improvement to enhance performance across a wider range of target curves. 

The FEA validation confirms the capability of the NN-generated multi-material designs to achieve the desired non linear mechanical behavior, representing a significant advancement in efficient design generation for additive manufacturing. This method has the potential to reduce design iteration times and minimize material waste, offering a powerful tool for accelerating innovation in the field. Moreover, our neural network-based approach extends beyond the capabilities of traditional topology optimization methods \cite{han2022multi}, which are typically gradient-based and well-suited for linear responses. By leveraging deep learning techniques, our method effectively addresses nonlinear phenomena such as large deformations and plasticity while able to generate multiple designs per condition, opening up new possibilities in metamaterial design.

Finally, it is worth noting that the validation framework outlined in this section resembles the actual design framework that can be deployed in practice. After providing a designer engineer with adequately trained diffusion and structural identifier models, their inference will almost instantly provide many multi-material design candidates for the desired nonlinear behavior. Based on their additive manufacturability, a few will be chosen for quick forward finite element analysis to find the one most closely matching the original nonlinear behavior, which will be 3D printed and further experimentally tested. Since the training of the generative AI  model on GPUs is usually done once, the trained learnable parameters (weights and biases) can be transferred to any low-end computing platform such as a laptop, and the multi-material design candidates can be inferred almost instantly for a given nonlinear response.

\section{Conclusions and future work}
\label{sec:conc}
The study presents a promising generative AI framework for multi-material lattice structure designs with tailored target responses. It demonstrates significant potential in fields requiring inverse-design problems, such as impact-resistant structures, additive manufacturing, topology optimization, and biomedical designs. The results show that the model can effectively produce multi-material designs that align with desired nonlinear stress-strain curves, bridging the gap between target performance and manufacturable outputs. While the framework shows considerable promise, areas remain for improvement to ensure consistent accuracy across all mechanical behaviors. We propose the following approaches to enhance the model's reliability and generalizability. First, we suggest to expand the dataset size. It could be advantageous to incorporate a wider range of data that covers more mechanical behaviors and material choices to enhance the generalizability of NNs. Second, one could implement enhanced algorithms for the loss functions in both the diffusion model and the structure identifier, capturing the data to get refined predictions in the inference phase. 
% Third, an improved noise reduction approach to predict designs from solution fields. In this work, we used two UNets to obtain the structure. On the other hand, advanced noise reduction algorithms could be applied to the diffusion output fields to get the multi-material structure quickly. 
Third, further improving the fidelity of the input finite element simulations by considering feature-size-dependent material properties induced by the additive manufacturing process employed to manufacture the multi-material structures \citep{he2023size,chadha2024exploring}. 
Lastly, integrate uncertainty quantification methods. Allowing the framework to predict the uncertainty of its creation by providing confidence intervals for the mechanical properties will enhance decision-making by allowing designers to assess the reliability of the generated designs, identify potential risks, and make informed choices regarding material use and design modifications.

In conclusion, this work, for the first time, demonstrates the effectiveness of a generative AI-based approach for creating multi-material designs with desired complex nonlinear mechanical behavior originating from large deformation and plasticity, and offering a significant advancement in computational design for additive and other advanced manufacturing processes. Our framework shows promise in revolutionizing inverse-design problems across various fields. Future efforts will address current limitations and extend the framework to more complex design scenarios, further pushing the boundaries of innovation in mechanics and materials science and engineering.

\section*{Replication of results}
The data and source code supporting this study will be available in a public GitHub repository.

\section*{Conflict of interest}
The authors declare that they have no conflict of interest.

\section*{Acknowledgements}
The authors would like to thank the National Center for Supercomputing Applications (NCSA) at the University of Illinois, and particularly its Research Computing Directorate, the Industry Program, and the Center for Artificial Intelligence Innovation (CAII) for their support and hardware and software resources. This research is a part of the Delta research computing project, which is supported by the National Science Foundation (award OCI 2005572) and the State of Illinois, as well as the Illinois Computes program supported by the University of Illinois Urbana-Champaign and the University of Illinois System. We also extend our gratitude to Jan-Hendrik Bastek and Dennis M. Kochmann for their original work on the video diffusion model, and for their valuable insights provided through email correspondence.

\section*{CRediT author contributions}
\textbf{Jaewan Park}: Conceptualization, Methodology, Software, Formal analysis, Investigation, Writing - Original Draft.
\textbf{Shashank Kushwaha}: Conceptualization, Methodology, Software, Formal analysis, Investigation, Writing - Original Draft.
\textbf{Junyan He}: Investigation, Supervision, Writing - Review \& Editing.
\textbf{Seid Koric}: Conceptualization, Methodology, Supervision, Resources, Writing - Original Draft, Funding Acquisition.
\textbf{Qibang Liu}: Investigation, Writing - Review \& Editing.
\textbf{Iwona Jasiuk}: Supervision, Writing - Review \& Editing.
\textbf{Diab Abueidda}: Conceptualization, Methodology, Supervision, Writing - Review \& Editing.

\bibliographystyle{unsrtnat}
\setlength{\bibsep}{0.0pt}
{\scriptsize \bibliography{References.bib} }

\section*{Appendix}
\label{sec:appendix}

\subsection*{Material properties}
\begin{figure}[h!] 
    \centering
    \includegraphics[width=0.9\textwidth]{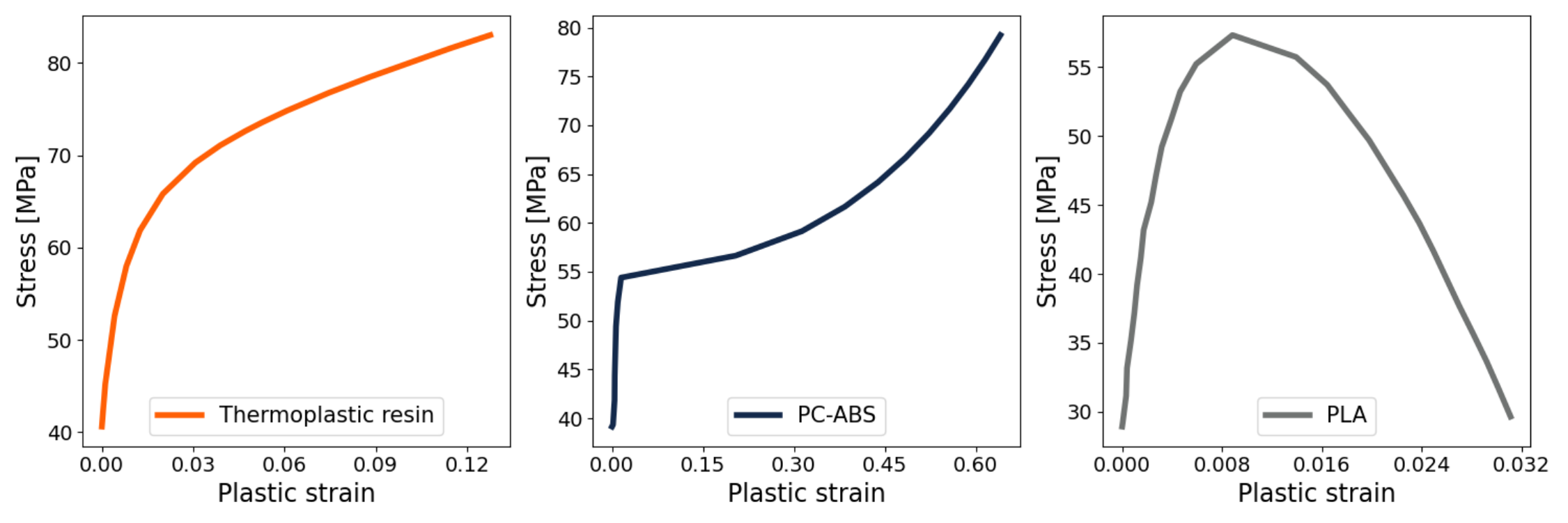}
    \caption{Stress and plastic stain plots for each material used in the work.}
    \label{fig:Material_plastic}
\end{figure}

The materials used in this work are as follows: thermoplastic acetal homopolymer resin, PC-ABS, and PLA. In the elastic regime, the thermoplastic acetal homopolymer resin has a Young's modulus of 2306 MPa and a Poisson's ratio of 0.35. For PC-ABS, the Young's modulus was set to 2500 MPa and a Poisson's ratio of 0.35. Lastly, for PLA, the Young's modulus was set to 2300 MPa and the Poisson's ratio was set to 0.35. The plastic regime for each material was plotted as shown in \fref{fig:Material_plastic}.

\end{document}